\documentclass[a4paper,fleqn]{cas-sc}

\usepackage[numbers]{natbib}

\def\tsc#1{\csdef{#1}{\textsc{\lowercase{#1}}\xspace}}
\tsc{WGM}
\tsc{QE}
\tsc{EP}
\tsc{PMS}
\tsc{BEC}
\tsc{DE}

\begin{document}
\let\WriteBookmarks\relax
\def\floatpagepagefraction{1}
\def\textpagefraction{.001}

\shorttitle{GRINN Framework}

\shortauthors{S Auddy et~al.}

\title [mode = title]{GRINN: A Physics-Informed Neural Network for solving hydrodynamic systems in the presence of self-gravity}                      



%
\author[ndAME]{Sayantan Auddy}[orcid=0000-0003-3784-8913]\corref{corxh}
\cortext[corxh]{Corresponding author}
\ead{sayantanauddy21@gmail.com,sayantan.auddy@jpl.nasa.gov}
\author[2,3]{Ramit Dey}[orcid=0000-0001-5717-1589]
\author[ndAME]{Neal J.\ Turner}[orcid=0000-0001-8292-1943]
\author[2,4]{Shantanu Basu}
[orcid=0000-0003-0855-350X]
\address[ndAME]{Jet Propulsion Laboratory, California Institute of Technology, Pasadena, CA 91109, USA}
\address[2]{Department of Physics \& Astronomy, University of Western Ontario, London, ON, N6A 3K7, Canada}
\address[3]{Perimeter Institute For Theoretical Physics, 31 Caroline St N, Waterloo, ON, Canada}
\address[4]{Institute for Earth \& Space Exploration, University of Western Ontario, London, ON, N6A 5B7, Canada}

\begin{abstract}
Modeling self-gravitating gas flows is essential to answering many fundamental questions in astrophysics. This spans many topics including planet-forming disks, star-forming clouds, galaxy formation, and the development of large-scale structures in the Universe. 
However, the nonlinear interaction between gravity and fluid dynamics offers a formidable challenge to solving the resulting time-dependent partial differential equations (PDEs) in three dimensions (3D). 
By leveraging the universal approximation capabilities of a neural network within a mesh-free framework, physics informed neural networks (PINNs) offer a new way of addressing this challenge. 
We introduce the gravity-informed neural network (GRINN), a PINN-based code, to simulate 3D self-gravitating hydrodynamic systems. 
Here, we specifically study gravitational instability and wave propagation in an isothermal gas. Our results match a linear analytic solution to within 1\% in the linear regime and a conventional grid code solution to within 5\% as the disturbance grows into the nonlinear regime. 
We find that the computation time of the GRINN does not scale with the number of dimensions.
This is in contrast to the scaling of the grid-based code for the hydrodynamic and self-gravity calculations as the number of dimensions is increased. 
Our results show that the GRINN computation time is longer than the grid code in one- and two- dimensional calculations but is an order of magnitude lesser than the grid code in 3D with similar accuracy.  Physics-informed neural networks like GRINN thus show promise for advancing our ability to model 3D astrophysical flows.
\end{abstract}



\begin{keywords}
 hydrodynamics \sep star formation \sep gravitational instability \sep neural networks \sep machine learning \sep PINN \sep PDEs
\end{keywords}

\maketitle

\section{Introduction}
The field of scientiﬁc machine learning (SciML), lying at the junction of data-based modeling and physics, has shown substantial potential for solving problems across science and engineering domains \cite{https://doi.org/10.1002/gamm.202100006,BERG201828,cuoco2020enhancing,george2018deep,2021ApJ...920....3A, 2022ApJ...936...93A}. Many phenomena involving space and/or time variations are modeled using partial differential equations (PDEs). Efforts to solve such PDEs have led to the development of scientific computing techniques and numerical approaches like finite element (FE) \cite{bams/1183504922,10.1115/1.4009129}, finite difference (FD) \cite{iserles_2008}, and finite volume (FV) \cite{Eymard2000FiniteVM} methods. While these traditional solvers are powerful, they often require substantial computational resources or some restrictive assumptions, particularly for three-dimensional, multiscale problems. In particular, their inability to resolve a wide enough range of length or time scales has thwarted the current progress in many areas of astrophysics, including simulations of star and galaxy formation.
In recent years, physics-informed neural networks (PINNs) have emerged as an exciting prospect in applying neural networks to solve both ordinary and partial differential equations \cite{RAISSI2019686,cuomo2022scientific}. This has provided an alternative cohesive framework for   solving forward/inverse problems \cite{RAISSI2019686, chen2020physics,2021JCoPh.42509913Y} and surrogate modeling \cite{zhu2019physics} driven by PDEs. PINNs explicitly incorporate the dynamical equations during the training process, thus ensuring that all constraints and conservation laws governing the system are satisfied. This allows the use of PINNs for accurate modeling of various phenomena in fluid dynamics \cite{cai2021physics,en16052343}, structural mechanics \cite{zhang2022analyses,haghighat2021physics}, and electromagnetism \cite{9999971,noakoasteen2020physics}. 

The self-gravity of (dark or visible) matter is the essential force that drives the dynamics and development of the cosmic web of structures \cite{springel2006}, the formation of galaxies \cite{2020NatRP...2...42V} and stars \cite{shu1987}, as well as the instabilities within protoplanetary disks \cite{vor06,2016ARA&A..54..271K}. 
Our goal is to take the first steps toward using PINNs to study the 
evolution of self-gravitating hydrodynamic systems. Here we apply this framework to the study of waves and instabilities within interstellar molecular gas clouds.
These are the sites of current-day star and planet formation.  Interstellar molecular gas is typically modeled 
using a set of gravito-hydrodynamic (GHD) equations. Disturbances that are large enough can become unstable due to self-gravity, resulting in local collapse (Jeans instability) \cite{1902RSPTA.199....1J}. This leads to the creation of pockets of localized high density within the fluctuating background density.
The nonlinear influence of self-gravity leads to a wide dynamic range in the density of the gas. 
Solving such a system analytically requires imposing various limiting and often drastic assumptions. Alternatively, one can apply numerical methods to more complex equations using approximations like the FD method \cite{iserles_2008}. However, with these methods, the large dynamic range of variation makes it challenging to retain numerical resolution, particularly in high-density regions. 
Even if such high-density regions are resolved, the required time step can be drastically reduced due to stability or accuracy criteria. 
This makes the modeling of long-term evolution very challenging as well as computationally expensive. 
For example, simulations of galaxy formation are typically limited by the need to model star formation and stellar feedback as sub-grid processes \cite{2017ARA&A..55...59N}. Furthermore, three-dimensional simulations of star-forming collapse that resolve the inner protostar and disk region 
are severely constrained. Even the most advanced codes can reach only up to $\sim 10^3$ yr past protostar formation \cite{2019ApJ...876..149M}, whereas observed protostars are in the age range $~10^4-10^6$ yr.
While still in the early stages of development, PINNs provide an alternate pathway that may eventually address some of these computational challenges.

In this paper, we develop a PINN-based PDE solver for a three-dimensional (3D) GHD system consisting of a set of coupled time-dependent PDEs.
PINNs provide a mesh-free framework where the resolution of the GHD simulations is not limited by the finite spacing of a grid \cite{cuomo2022scientific}. 
In the FD approach, increasing dimensionality increases memory requirements geometrically, and increased resolution requires smaller time stepping in order to maintain accuracy or stability. Together these dramatically enhance the computation cost for 3D calculations with a high enough resolution to achieve sufficient fidelity. However, PINNs can adapt to varying resolutions in a more flexible way and will not necessarily follow these constraints.
Owing to the universal approximation capabilities of neural networks \cite{Hornik1989MultilayerFN}, PINNs have the potential to capture the nonlinear interaction between fluid dynamics and gravity with high accuracy. Moreover, the traditional FD-based approach can suffer from accuracy and stability issues when applied to nonlinear systems that contain small-scale structures and discontinuities (e.g., a shock front).
 PINN-based models have particularly demonstrated their ability in dealing with PDEs whose solutions can develop shocks. \cite{coutinho2023physics,MAO2020112789}.

The paper is organized as follows, we discuss the hydrodynamic system of interest in \S\ref{sec:hydrosystem} and introduce the PINN algorithm in \S\ref{sec:PINN-architecture}. In  \S\ref{sec:results} we give three case studies demonstrating the application of GRINN and compare the results with linear theory and FD methods. We discuss the limitations and potential extensions in \S\ref{sec:discussion}  and summarize our findings in \S\ref{sec:conclusion}.




\section{Hydrodynamic System}\label{sec:hydrosystem}
We solve the system of isothermal self-gravitating hydrodynamics, which is of interest in the study of 
interstellar molecular clouds, the sites of current-day star formation. These are dark cold regions of molecular (mostly H$_2$) gas in which the self-gravity makes density perturbations unstable. This can trigger a local collapse resulting in the growth of dense cores within the gas clouds that collapse further to form stars. The initial conditions that trigger the collapse of molecular clouds to form stars have been studied for decades \cite{mestel1956,shu1987,mousch1999,mckee2007,tsuka2022}. The simplest case of interest is to investigate the growth of gravitational instability in an isothermal gas while accounting for thermal pressure and gravitational effects.  Since the radiative cooling time is shorter than the collapse time in the early stages, the gas is close to isothermal until well into the final collapse that produces a star.


The behavior of the interstellar gas is governed by a set of fundamental hydrodynamic equations. We solve the three-dimensional (3D) hydrodynamic equations including self-gravity to study the gravitational (or Jeans \citep{1902RSPTA.199....1J}) instability in star-forming molecular clouds. If $\rho$ and $\textbf{v}$ are the gas density and velocity respectively we can express the governing equations as 
\begin{align}\label{cont}
    &\partial_t \rho + \nabla \cdot (\rho\, \textbf{v}) =0\, , \\ \label{mom}
    &\rho \left[\partial_t \textbf{v} + (\textbf{v} \cdot \nabla)\textbf{v} \right] = - \nabla P + \rho\, \textbf{g} \, ,\\ \label{poi}
    &\nabla^2 \phi =  4 \pi G \rho  \, .
\end{align} \label{hydro}
Eqs. (\ref{cont} - \ref{poi}) are the equations of mass continuity, momentum, and self-gravity (the Poisson equation), respectively. The 
gravitational field $\textbf{g} = - \nabla \phi $, where $\phi$ is the gravitational potential given in Eq. (\ref{poi}) and $G$ is the gravitational constant. We close the above set of equations with the generalized 
polytropic relation $P = K \rho^{\gamma}$. We consider isothermal evolution ($\gamma = 1$) of an ideal gas in which $P = \rho kT/m$, where  $T$ is the temperature, $m$ is the mean mass of molecules, and hereafter we identify $c_s \equiv (kT/m)^{1/2}$ as the isothermal sound speed. This coupled set of equations has a diverse set of solutions depending on the initial and the boundary conditions of the system of interest. In this paper, we apply 
sinusoidal
perturbations to the background steady-state fluid and study the behavior under the influence of self-gravity. 

Solving Eqs. (\ref{cont} - \ref{poi}) under a linear approximation (Appendix \ref{LT}) gives a characteristic length scale for instability, the Jeans length $\lambda_{J} \equiv (\pi c_s^2/G \rho_0)^{1/2}$, where $\rho_{0}$ is the uniform background density. For any perturbation with wavelength $\lambda > \lambda_{J}$ the amplitude grows exponentially in time. The overdense region is unstable and goes into a runaway collapse. For $\lambda < \lambda_{J}$, the system is stable and exhibits local oscillatory behavior as well as wave propagation. 

Here we  consider both linear and nonlinear initial sinusoidal perturbations. Linear perturbations are small amplitude ($<0.1 \rho_0$) waves while nonlinear perturbations are large amplitude ($>0.1 \rho_0$) disturbances to the uniform background density $\rho_0$. 
This fluid system has known analytic solutions, letting us probe the effectiveness of the PINN architecture introduced in the next section for solving the GHD equations.






\begin{figure}
\centering
\includegraphics[scale=.18]{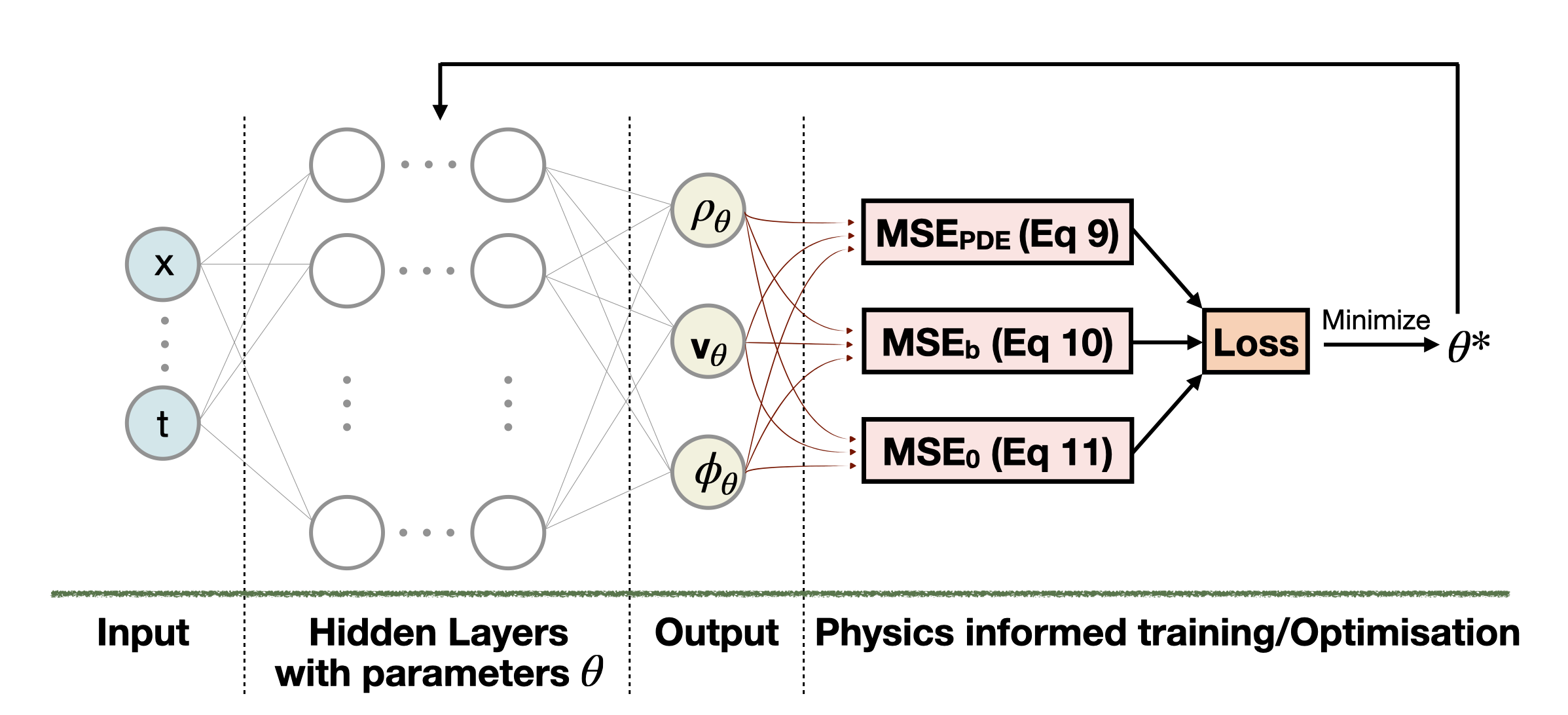}
\caption{ Schematic of the PINN-based GRINN workflow. The output of the fully connected neural network is $\rho_{\theta}, \textbf{v}_{\theta}, \phi_{\theta}$. These approximate solutions are used along with the PDEs governing the system to obtain the total loss function. During the training process, the parameters $\theta$ are optimized iteratively to obtain $\theta^*$. Using this the final solutions of the system are obtained.    }\label{fig:schematic}
\end{figure}

\section{PINNs Architecture}\label{sec:PINN-architecture} 


A PINN is a neural network configured to simulate a dynamical system governed by PDEs. This is done by approximating the PDE's solutions by training the neural network $\mathcal{N} (X;\theta)$ to minimize a loss function (or the residual).  Here $\mathcal{N}(X;\theta)$ has parameters $\theta$ (weights and biases of the neurons) and is defined over the space-time domain ($d$ spatial and one temporal dimension)  denoted by collocation points $X:= [\textbf{x},t]$. These collocation points are a set of points sampled from the domain of integration and are given as input to the model. Instead of directly solving the equations, the PINN's architecture solves the PDEs as a loss function optimization problem. The novel aspect of PINNs is the incorporation of a residual network that encodes the PDEs along with the boundary and initial conditions. In particular, the loss function is reinforced with a residual term originating from the governing equations and this acts as a penalty term that quantifies the deviations from the ground truth solution of the PDEs.  For the forward problem, we train  PINNs   by an unsupervised learning approach based solely on the physical equations as well as the boundary/initial conditions without the aid of any labeled data (i.e., no additional observational or simulation data are needed  as input). 

 In this section, we introduce GRINN, a PINN-based model used for simulating the hydrodynamic system in the presence of self-gravity. 
Consider a parametrized differential equation in a general form 
\begin{align}
    &\mathcal{D}\left[ \mathcal{N} (X;\theta), \Lambda \right] = f(X), \, X \in \Omega\, , \nonumber \\ 
    &\mathcal{B}\left[ \mathcal{N} (X;\theta) \right] = g(X), \, X \in \partial\Omega\, ,
\end{align}
defined on the domain $\Omega \in  \mathbb{R}^{d} $ with the boundary $\partial \Omega$. Here $\mathcal{D}$ is the nonlinear differential operator, $f(X)$ is a source term, $\Lambda$ is a vector of additional parameters of the differential operator, $\mathcal{B}$ is a set of boundary/initial conditions related to the problem and $g$ is the boundary function.

Our system of interest involves studying the dynamics of the self-gravitating gas. The governing Eqs. (\ref{cont} - \ref{poi}) are coupled to each other where the gas density $\rho$, velocity $\textbf{v}$, and the gravitational potential $\phi$ are dependent variables. We begin with a single neural network $ \mathcal{N}(X;\theta)$ and approximate the solutions $\rho$, $ \textbf{v}$, and $\phi $ of the PDEs as the outputs of the network,
 \begin{eqnarray}
\mathcal N(X;\theta)  &\simeq \rho_\theta(X), \textbf{v}^i_\theta(X),  \phi_\theta(X)\, . 
\end{eqnarray}
In this approach, the total loss is defined as the sum of the PDE residual, the boundary, and initial condition residuals at the collocation points $X$ for each of the output variables. The PDE's residuals associated with Eqs. (\ref{cont}-\ref{poi}) for a ($d+1$)-dimensional system are given as
\begin{align} \label{res}
    &R_{\rho}(X)=\partial_t \rho_{\theta}(X) + \sum_{i=1}^d\partial_{x_i}.(\rho_{\theta}(X)  \textbf{v}_{\theta}^i(X))\, ,
    \\ \label{res2}
    &R_{ \textbf{v}^i}(X)=\rho_{\theta}(X) \left[\partial_t \textbf{v}_{\theta}^i(X) + (\sum_{j=1}^{d}\textbf{v}_{\theta}^j  \partial_{x^j}) \textbf{v}_{\theta}^i \right] +  c^2_s\partial_{x^i}\rho_{\theta}(X) - \rho_{\theta}(X) \textbf{g}_{\theta}^i(X)\, ,
    \\ \label{res3}
    &R_{\phi}(X)=\sum_{i=1}^{d}\partial^2_{x^i} \phi_{\theta}(X) -  4 \pi G \rho_{\theta}(X)\, .
\end{align}
 The MSE corresponding to the residuals computed in Eq. (\ref{res} -\ref{res3}) for a set of randomly generated collocation points sampled from a uniform distribution in 3+1 dimensions $X^r_n:=[x^r_n,y^r_n,z^r_n,t^r_n]$ is given as
\begin{align} \label{mse_pde}
MSE_{PDE} (\theta)& =
\frac{1}{N_r}\sum_{n=1}^{N_r} \left|R_{\rho}(X^r_n) \right|^2 +
\sum_{j=1}^d\bigg(\frac{1}{N_r}\sum_{n=1}^{N_r} \left|R_{\textbf{v}^j}(X^r_n) \right|^2 \bigg)
+\frac{1}{N_r}\sum_{n=1}^{N_r} \left|R_{\phi}(X^r_n) \right|^2 .
\end{align}
where $N_r$ is the total number of collocation points in the space-time domain.  The MSE associated with the boundary and initial conditions for density $\rho$ , velocity  $\textbf{v}^i_{\theta}$ and gravitational potential $\phi_{\theta}$ at randomly generated collocation points $X_n^b:=[x^b_n,y^b_n,z^b_n,t^b_n]$ and $X_n^0:= [x^0_n,y^0_n,z^0_n,t^0_n]$ are
\begin{align} \label{mse_b}
    &MSE_{b}=\frac{1}{N_b}
\sum_{n=1}^{N_b} \left|\rho_{\theta}(X^b_n) - \rho_{b}(X^b_n)\right|^2
+\sum_{j=1}^d\left(\frac{1}{N_b}\sum_{n=1}^{N_b} \left|\textbf{v}^j_{\theta}(X^b_n) - \textbf{v}^j_{b}(X^b_n)\right|^2\right)
+\frac{1}{N_b}
\sum_{n=1}^{N_b} \left|\phi_{\theta}(X^b_n) - \phi_{b}(X^b_n)\right|^2\, ,
    \\
    \label{mse_0}
    &MSE_0=\frac{1}{N_0}
\sum_{n=1}^{N_0} \left|\rho_{\theta}(X^0_n) - \rho_{0}(X^0_n)\right|^2
+\sum_{j=1}^d\left(\frac{1}{N_0}\sum_{n=1}^{N_0} \left|\textbf{v}^j_{\theta}(X^0_n) - \textbf{v}^j_{0}(X^0_n)\right|^2\right)
+\frac{1}{N_0}
\sum_{n=1}^{N_0} \left|\phi_{\theta}(X^0_n) - \phi_{0}(X^0_n)\right|^2\, .
\end{align}
Here, $N_b$ and $N_0$ are the total number of collocation points on the boundary and at the initial time, respectively.
The summation over $j$ takes into account the velocity components ($\textbf{v}_{\theta}^x,\textbf{ v}_{\theta}^y,\textbf{ v}_{\theta}^z$) in the three spatial dimensions for a ($3+1$) dimensional system. The training objective is to minimize the total loss  $(MSE_{PDE}+MSE_{b}+MSE_0)$ originating from the residuals of the PDE and the boundary and initial conditions. This is done by optimizing the neural network parameters $\theta$ to yield
\begin{align} \label{total_loss}
    \theta^*={\rm argmin}\,(MSE_{PDE}+MSE_{b}+MSE_0)\, .
\end{align}
Once trained, the PINN model can accurately approximate the solutions of the PDEs. The training process (i.e.,  optimization of the network parameters) works iteratively, where the derivatives of the outputs are taken with respect to the network parameters in order to compute the MSEs (as given in Eqs. (\ref{mse_pde}-\ref{mse_0})) and adjust $\theta$ in each step to minimize the MSE.  To ensure the differentiability of the  output, which is the solution approximated by the neural network, a smooth activation function is used (e.g., tanh, sigmoid, sine).
PINNs exploit the auto-differentiation \cite{JMLR:v18:17-468} technique for estimating the derivative of the outputs and eventually obtaining the PDE solutions. Furthermore, numerical methods such as FD show inefficiency when applied to complicated nonlinear functions.  Automatic differentiation overcomes these  limitations and shows no approximation (truncation) error as well.  This enables the network to approximate any PDE along with the given boundary conditions without numerically discretizing the equations, thus not requiring a mesh.

We adopt {\tt\string  DeepXDE} \cite{lu2021deepxde} to design our PINN model GRINN. It is a fully connected neural network with  $3$ hidden layers having $32$ neurons in each layer. Fig. \ref{fig:schematic} illustrates the schematic of the GRINN architecture. Here, the $sine$ activation function is implemented for each neuron. The smoothness of the $sine$ activation function enables the network to achieve more stability as well as better training efficiency compared to step functions like a sigmoid \cite{parascandolo2017taming,sitzmann2020implicit}. We randomly sample $N_r=47000$ collocation points over the given space-time domain. This acts as an input to the network which computes the residuals in Eqs. (\ref{res} - \ref{res3}) at each of these collocation points. Additionally, $N_b=N_0=6300$ points are sampled for the spatial boundary and at the initial time slice respectively.  The parameters $\theta$  are initialized using a truncated normal distribution (He-Normal).  
For the first 2000 epochs, we implement an adaptive stochastic gradient descent-based optimizer ({\tt\string ADAM})  with a learning rate of $1e-3$ to pre-train the network. Pre-training the network minimizes the chance of the optimization getting stuck at a local minimum. This is followed by training the network with the second-order quasi-Newton {\tt\string L-BFGS} optimizer \cite{liu1989limited} to converge on the global minima.    In this work, we choose the hyperparameters by a trial and error approach which minimizes computational cost without compromising accuracy.

\begin{figure*}[ht!]
\centering
\includegraphics[scale=.35]{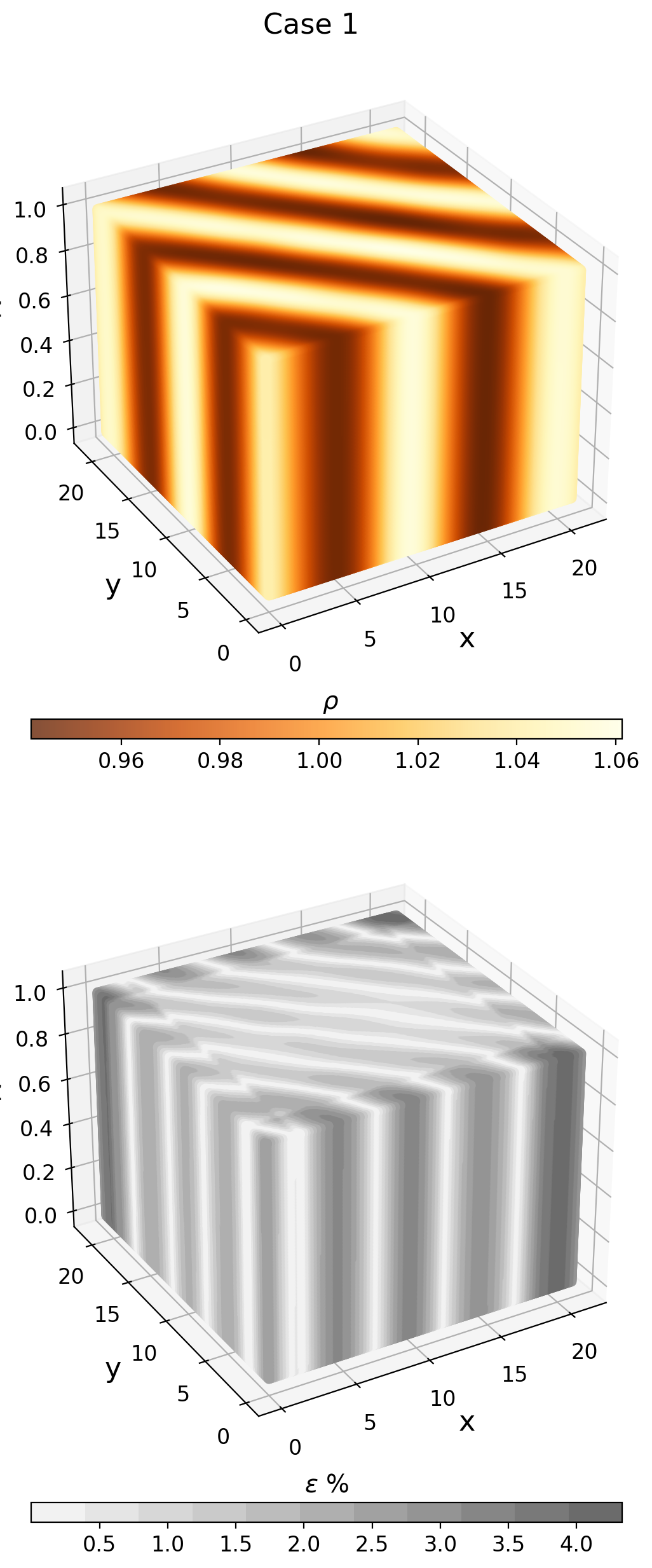}
\includegraphics[scale=.35]{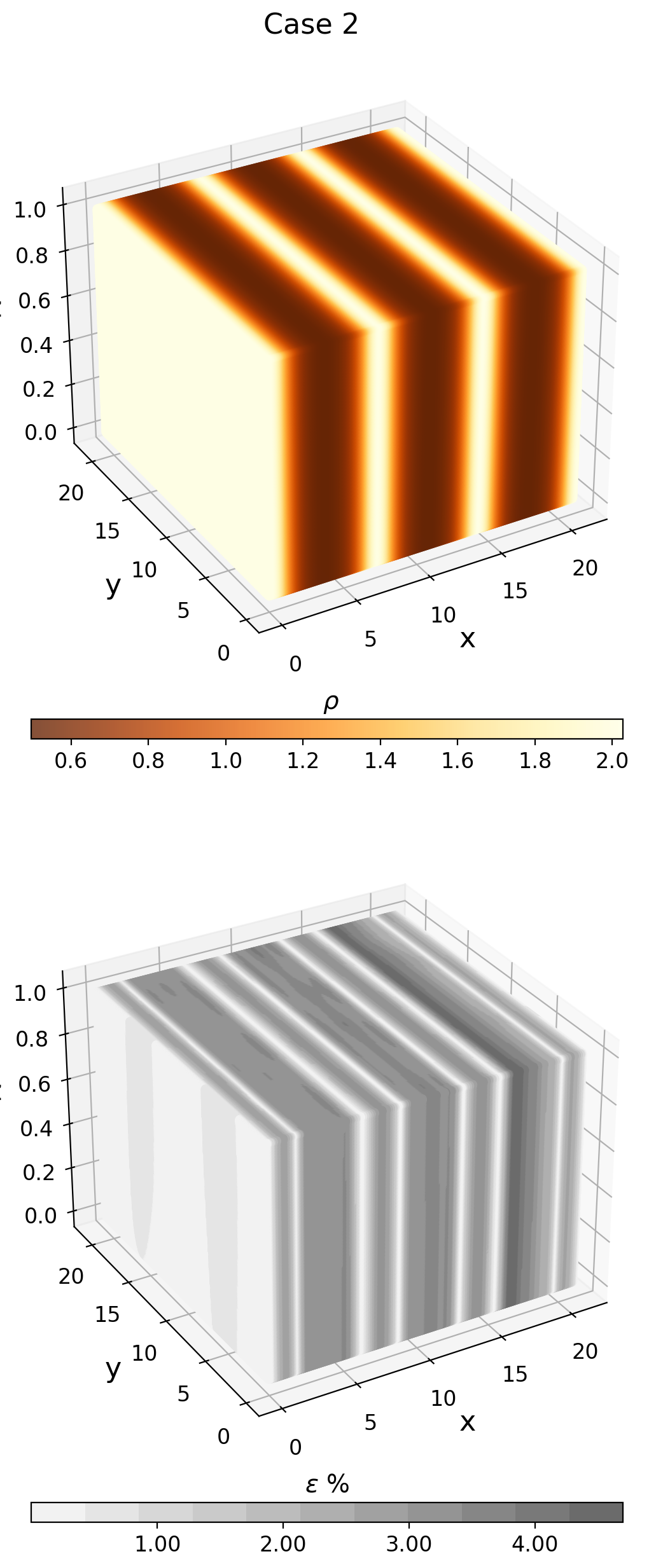}
\includegraphics[scale=.35]{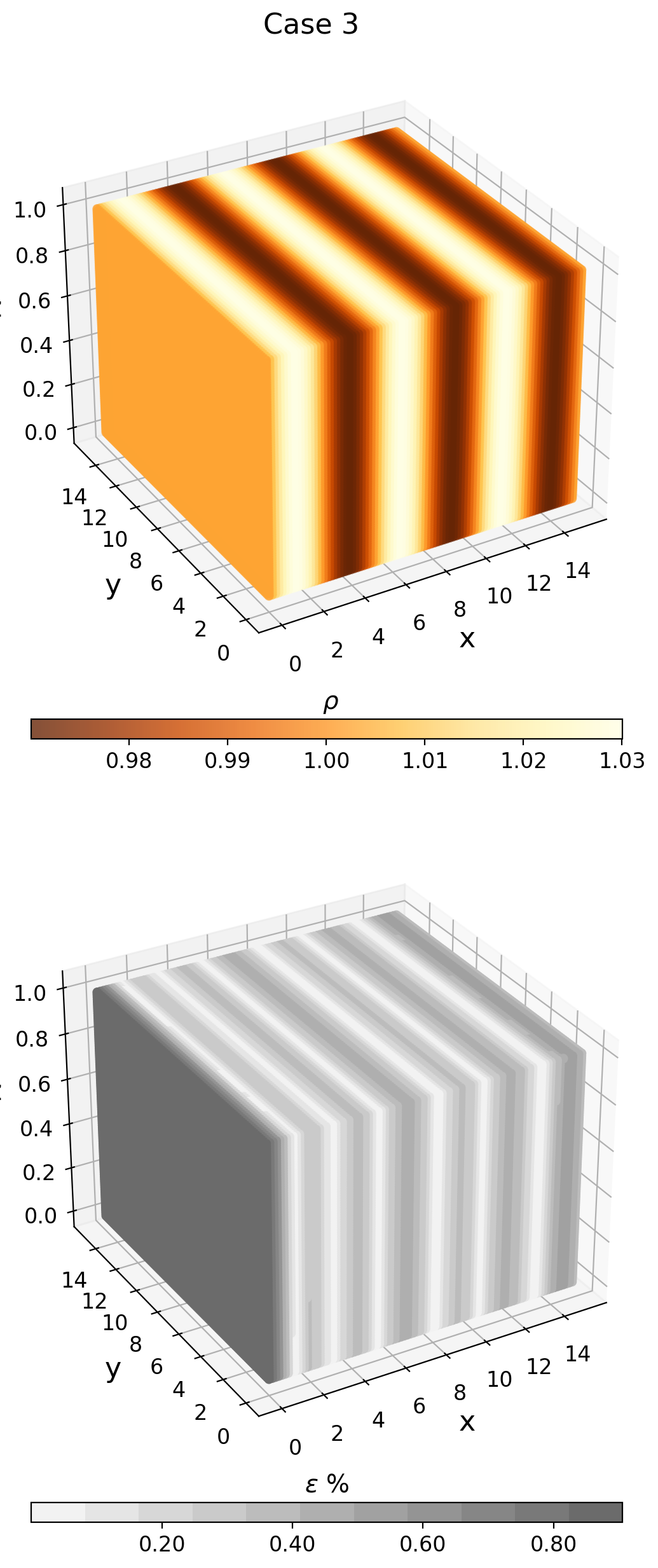}
\caption{Top: GRINN density solutions for a 3D self-gravitating hydrodynamic system (Eqs. \ref{cont}-\ref{poi}) for different initial conditions (three distinct cases) as discussed in \S \ref{sec:results} at $t=2$. Bottom: The relative mismatch (i.e., $\epsilon$ given in Eq.(\ref {relmisfit})) between the GRINN and standard FD solutions.}\label{fig:c1_3d}
\end{figure*}

\section{Results}\label{sec:results}
In this section, we demonstrate the GRINN approach's effectiveness in solving Eqs.~(\ref{cont} - \ref{poi}), which govern the gas dynamics in molecular clouds under the influence of gravity and thermal pressure. GRINN estimates the density, velocity, and gravitational fields as functions of space and time over a specified domain and time period.  We examine three distinct test problems, in each case comparing the results with solutions obtained using the standard FD method as well as an analytic solution using the linear theory (LT) solution set out in Appendix \ref{LT}.  For the FD solution we use the Lax method \cite{lax54} since it is an efficient explicit scheme with second-order truncation error that maintains monotonicity and can achieve numerical stability, while being relatively simple to implement.  We quantitatively assess the GRINN performance using the mismatch $\epsilon$ of the density, velocity, or gravitational field relative to the FD or LT solution, defined by
\begin{align}\label{relmisfit}
\epsilon =  \frac{2|(\rho, \textbf{v}, \phi)_{\rm GRINN} - (\rho, \textbf {v}, \phi)_{\rm FD/LT}|}{{ (\rho, \textbf{v}, \phi)_{\rm GRINN} + (\rho, \textbf{v}, \phi)_{\rm FD/LT}} } \times 100\, .
\end{align}

In \S\ref{sec:case1} we present the evolution of a small-amplitude growing disturbance that is unstable. In \S\ref{sec:case2} we study the effect of a large-amplitude  nonlinear disturbance that grows rapidly. Finally, in \S\ref{sec:case3} a small-amplitude stable wave is shown.



\begin{figure}[ht!]
\centering
\includegraphics[scale=.28]{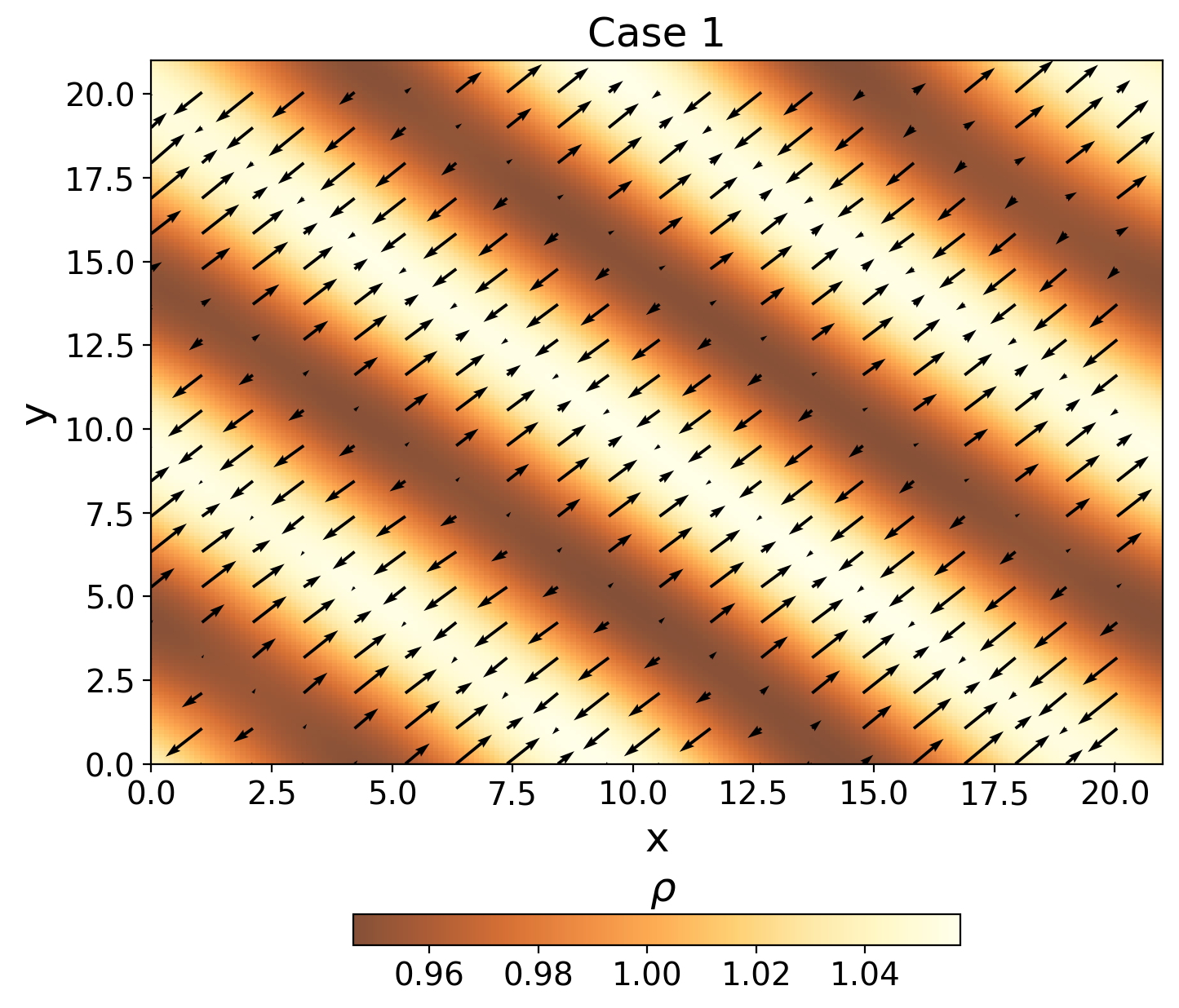}
\includegraphics[scale=.28]{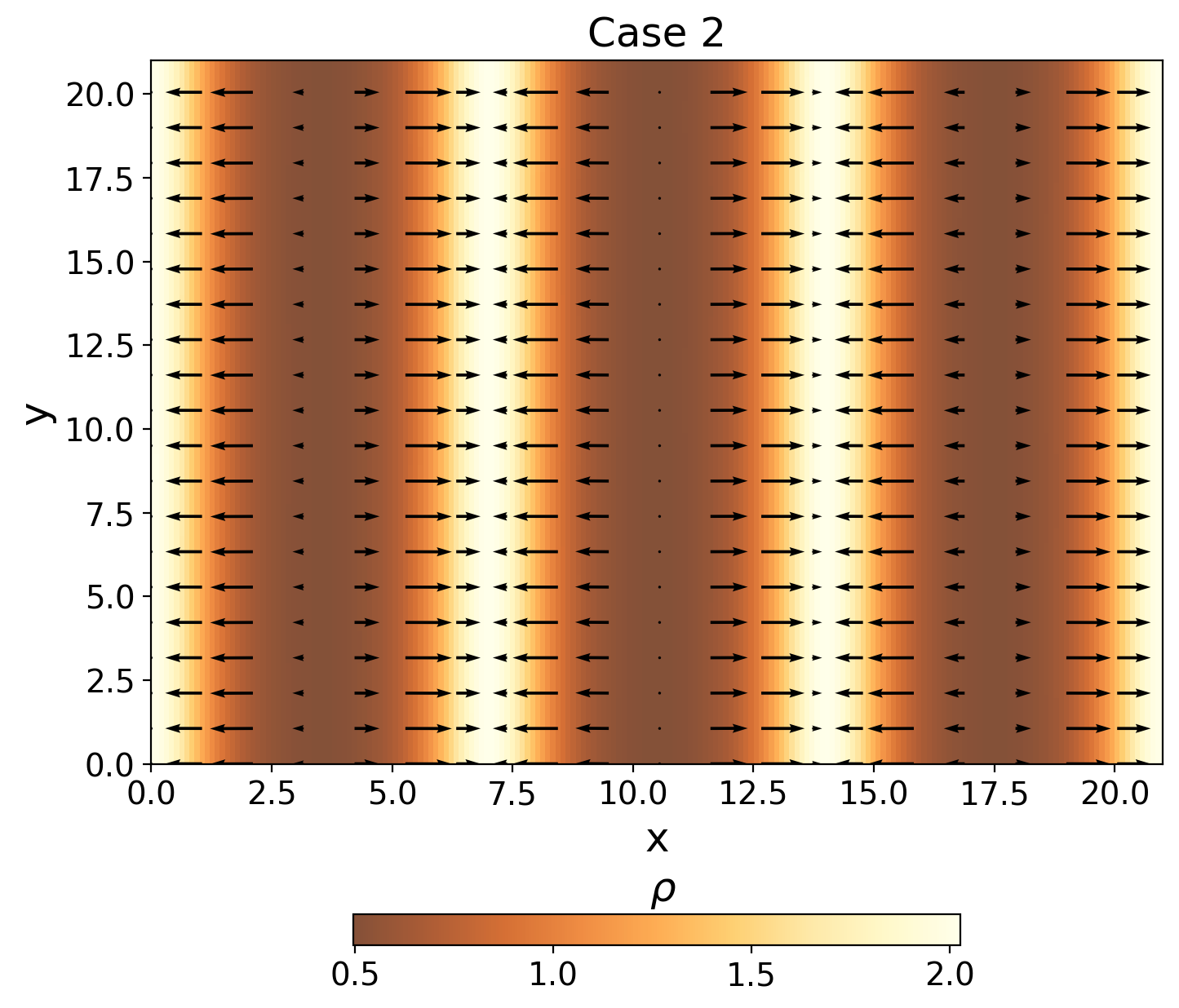}
\includegraphics[scale=.28]{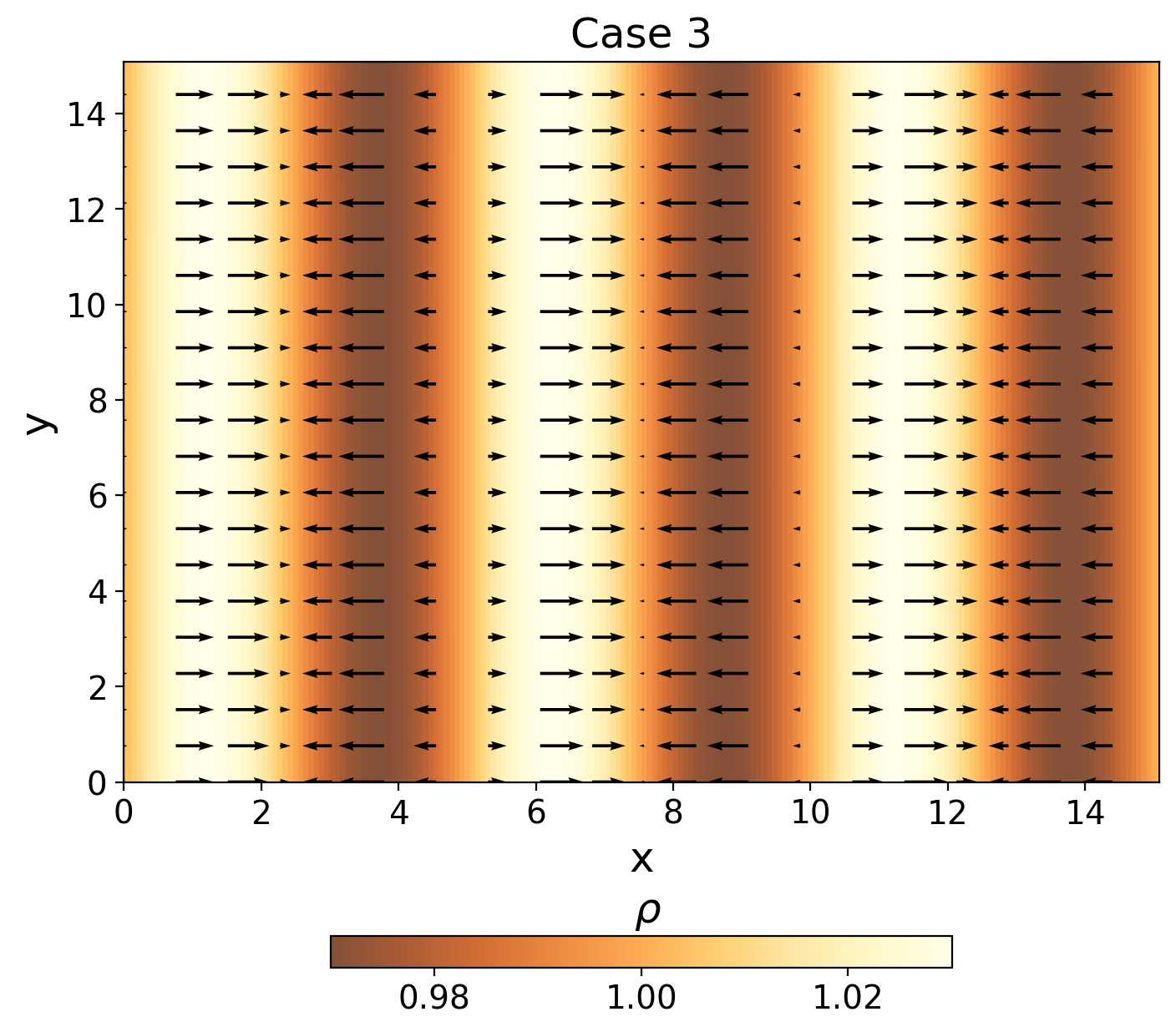}
\caption{Cross-sections through the GRINN density solutions for the three test cases in the $x-y$-plane at $z=0.6$ and $t=2$.  Velocity vectors are overplotted.\label{fig:c_plot}}
\end{figure}




\begin{figure*}[ht!]
\centering
\includegraphics[scale=.52]{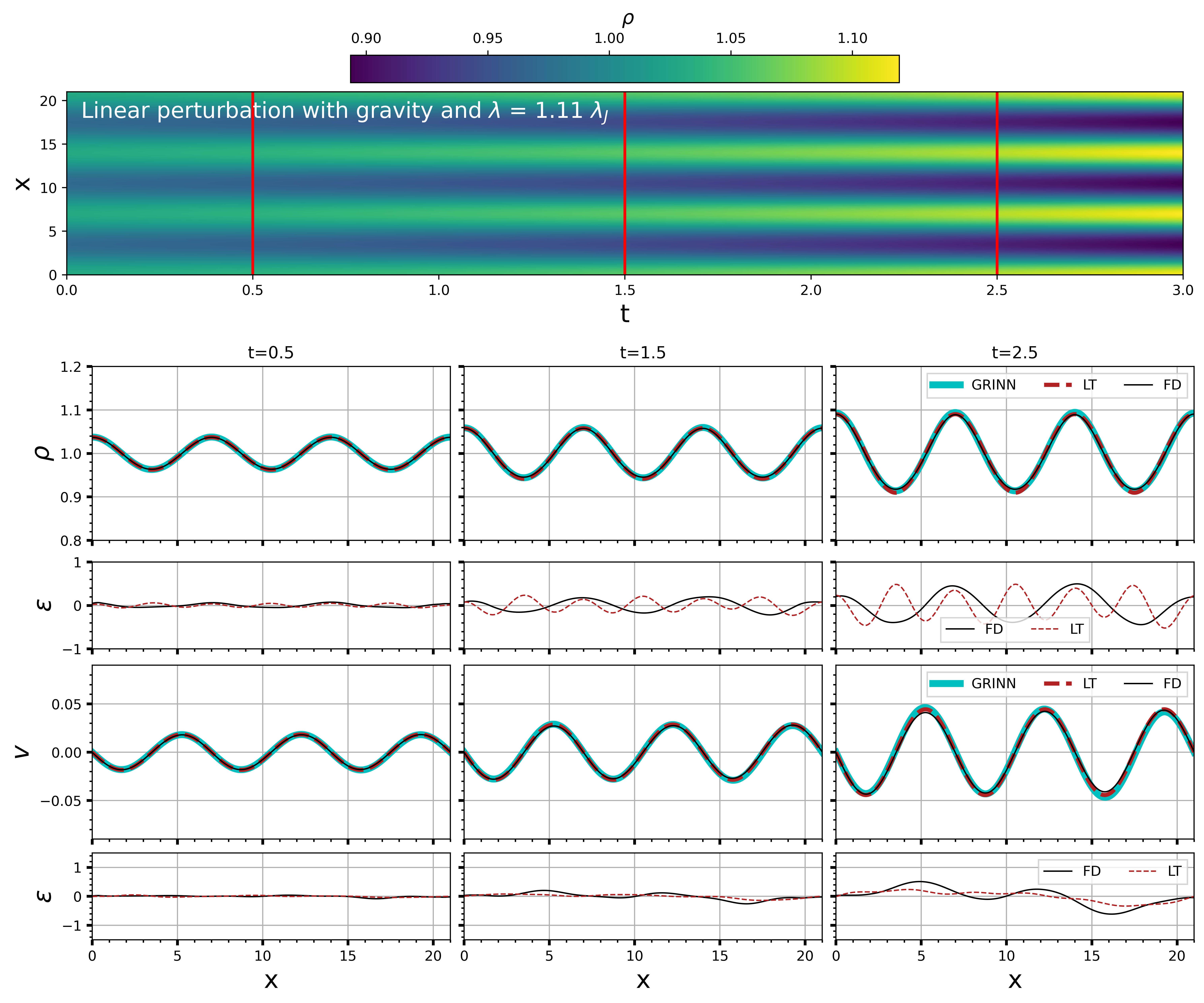}
\caption{The GRINN solution for case~1.
The top-row image shows the growth in the gas density over time due to Jeans instability from GRINN.  The next row of panels shows the density profiles from the GRINN, FD, and LT methods at the three times indicated by the red lines in the top-row image.  The third row shows the mismatch of the GRINN solution with respect to the FD and LT solutions. The fourth and fifth rows of panels give the corresponding velocity profiles and mismatches. The GRINN solution agrees closely with the FD and LT solutions, and the mismatches are smaller than 1\%. 
}\label{fig:jean-linear-gravity}
\end{figure*}


\subsection{Case 1: Growing linear perturbation with $\lambda>\lambda_{J}$ \label{sec:case1}}
We model the growth of the Jeans instability in a self-gravitating cloud of molecular hydrogen gas filling a cube defined in Cartesian coordinates.  GRINN is trained on the governing Eqs. 
 (\ref{cont} - \ref{poi}) for initial conditions with a perturbation in density $\rho$ and 1D velocity amplitude $v$ that corresponds to an individual mode under the LT:
\begin{align}\label{ic1}
    &\rho(\textbf{x},t=0) = \rho_o + \rho_{1a} \cos (\textbf{k}\cdot \textbf{x})\, ,  \nonumber \\
    & v (\textbf{x},t=0) = v_{1a} \sin (\textbf{k}\cdot \textbf{x})\, ,
\end{align}
where $v_{1a} = -\frac{\alpha}{ |\textbf{k}|}\frac{\rho_{1a}}{\rho_0}$, $\textbf{k}$ is the wavevector, and the mode's growth rate is $\alpha=\sqrt{4\pi G\rho_0-c_s^2k^2}$, as derived in Appendix~\ref{LT}.

Henceforth, we work in a set of units in which $c_s = 4\pi G = \rho_0 = 1$. This means that all densities and speeds are normalized to $\rho_0$ and $c_s$, respectively. The units of time and length are $1/\sqrt{ 4\pi G\rho_0}$ and $c_s/\sqrt{ 4\pi G\rho_0}$, respectively. 

The case~1 initial condition is a linear wave of amplitude $\rho_{1a} = 0.03$ with fronts inclined $45^{\circ}$ in the $x-y$ plane using $k_x=k_y=k/\sqrt{2},k_z=0$.  We set the wavelength $\lambda = 1.11 \lambda_{J}$ (where $\lambda_{J}=2\pi$ in our units).  The computational domain spans three of these wavelengths.  Periodic boundary conditions (BCs) are enforced in all three directions for $\rho$ and $\textbf{v}$.  For example, along the $x$-direction,
\begin{align}\label{bc1}
    &\rho(x=0,y,z,t)= \rho(x_m,y,z,t)\, , \nonumber \\
    & \textbf{v}(x=0,y,z,t)= \textbf{v} (x_m,y,z,t)\, ,
\end{align}
where $x_m$ is the domain length along the $x$-direction.  Since Eq.~(\ref{poi}) is a second-order differential equation, we enforce periodicity for both the gravitational potential $\phi$ and its derivative $\phi^\prime$ perpendicular to the boundary, so that
\begin{align}\label{bc_phi1}
    \phi (x=0,y,z,t)&= \phi(x_m,y,z,t)\, ,\nonumber \\
    \phi^\prime (x=0,y,z,t)&= \phi^\prime (x_m,y,z,t)\, .
\end{align}
The periodic BC ensures that gas leaving the domain through any boundary re-enters from the opposite side. 

The training process involves optimizing the model parameters $\theta$ by minimizing the total loss in Eq.~\eqref{total_loss} for the PDEs Eqs.~(\ref{cont} - \ref{poi}), initial conditions Eq.~(\ref{ic1}), and BCs Eqs.~(\ref{bc1} - \ref{bc_phi1}). The minimization yields 3D solutions for the gas density, velocity, and gravitational field as functions of time over the selected interval.  The density result for case~1 at time $t=2$ is in Fig.~\ref{fig:c1_3d} on the left.  The lower panel shows the mismatch $\epsilon<2.5\%$ between the GRINN and FD solutions.  Fig.~\ref{fig:c_plot} is an $x-y$ cross-section through the domain at $z=0.6$ and $t=2$.  For the FD calculations in 3D, we choose the Courant number $\nu = 0.5 $ and  the number of grid points $N^3=300^3$. At the final time $t=3$, the Jeans length at the maximum density is resolved with 80 grid points in each direction. Arrows are velocity vectors depicting the direction of flow.  For case~1 the gas flows into the overdense regions, further increasing the density with time as the unstable mode grows.

For further insight into case~1, we apply a simpler initial perturbation along the $x$-axis, setting $k_x=k$ and $k_y=k_z=0$ in Eq.~\eqref{ic1}.  Cuts along the $x$-direction at $y=0.6$ and $z=0.6$ are in Fig.~\ref{fig:jean-linear-gravity}.  The top-row image shows the gas density found by GRINN versus $x$ and $t$.  The density grows over time due to the Jeans instability.  In the panels below, we show cuts through the solution at the three times marked by red lines in the top-row image: $t=0.5$, $1.5$, and $2.5$, comparing with the corresponding FD and LT solutions. For the FD calculations in 1D, we increase the number of grid points to $N=1000$ with Courant number $\nu = 0.5$ for better accuracy.  Eq.~(\ref{alpha}) indicates the instability should grow with an $e$-folding time $\tau=2.3$, which is approximately borne out as the denser regions become more dense with increasing time.  The gas flows out of the less-dense regions at increasing speed and into the density peaks. The gas velocity variation shown in the second row from the bottom is $90^\circ$ out of phase with the density so that zero velocity coincides with the maximum density.  The GRINN solutions for both $\rho$ and $\textbf{v}$ are consistent with the LT and FD solutions with relative mismatch $<1.0 \%$ at all times shown.

\begin{figure}
\centering
\includegraphics[scale=.52]{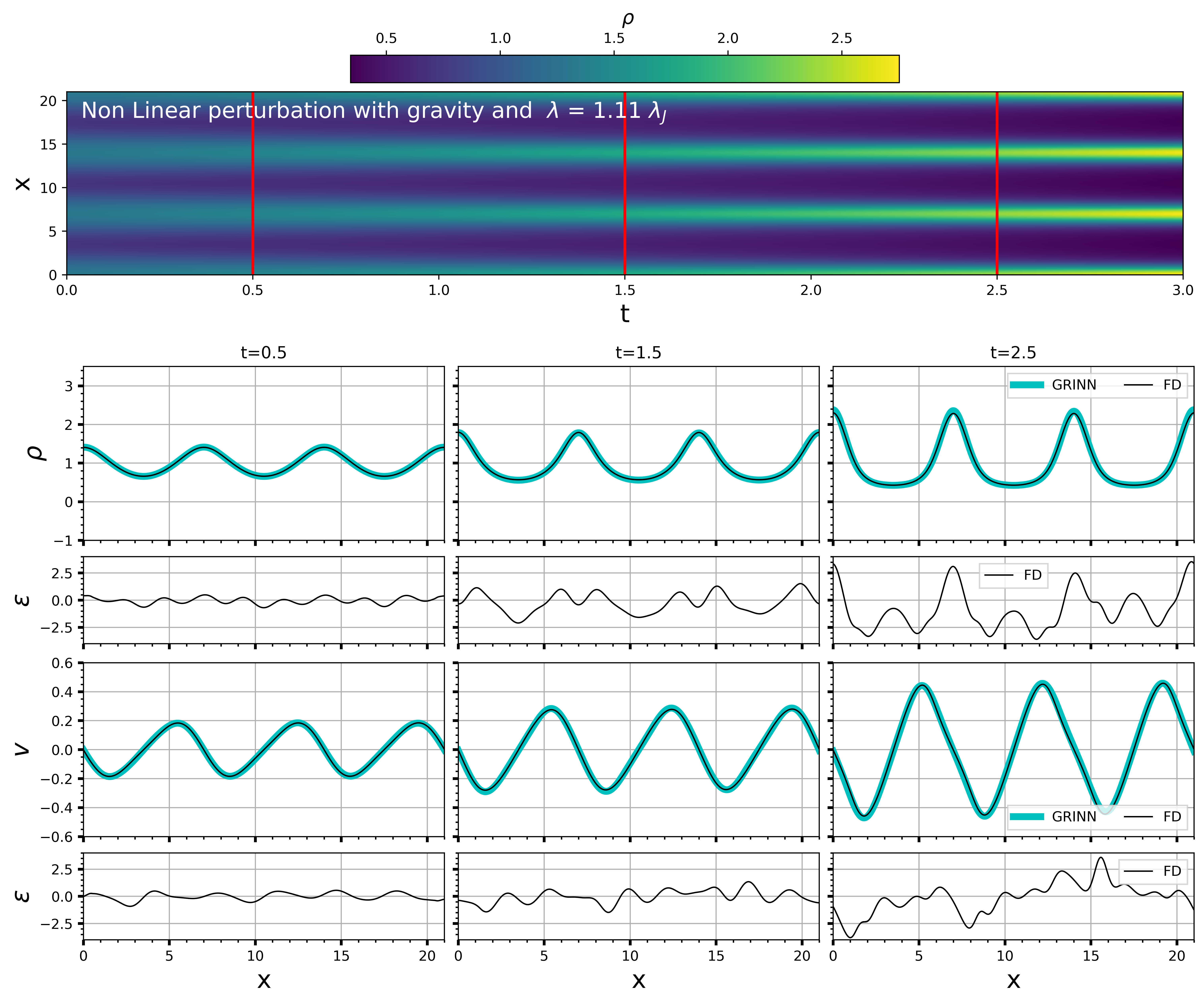}
\caption{Same as Fig~\ref{fig:jean-linear-gravity} but for case~2, with larger initial density perturbation of amplitude $\rho_{1a} = 0.3$. 
}\label{fig:jean-nonlinear-gravity}
\end{figure}

\subsection{Case 2: Growing nonlinear perturbation with $\lambda>\lambda_{J}$ \label{sec:case2}}
We next examine how GRINN performs on a large-amplitude, nonlinear initial perturbation, consisting of a wave with amplitude $\rho_{1a} = 0.3$ and wavelength $\lambda = 1.11 \lambda_{J}$.  
We orient the wavevector along the $x$-axis, with $k_x=k$ and $k_y=k_z=0$ in Eq.~\eqref{ic1}.

The case~2 results are shown in Fig.~\ref{fig:c1_3d} middle column.  The upper plot has the density profile from GRINN at $t=2$.  The amplitude is greater than in case~1 because of the greater initial perturbation, as indicated by the differing color scales across the figure's columns.
The corresponding 2D slice showing the flow directions appears in the middle panel of Fig.~\ref{fig:c_plot}.  As in case~1, the flow is into the density peaks.  The GRINN and FD results agree to within $\epsilon \lesssim 5\%$ for $t<2$.  However, the mismatch grows along with the nonlinearity at later times.
Fig.~\ref{fig:jean-nonlinear-gravity} shows 1D cuts through the growing density and velocity at $y=z=0.6$. For the FD calculations in 1D,  we adjust the Courant number to $\nu = 0.6 $ and increase the number of grid points  to $N=2000$ for improved accuracy. Since the linear theory is invalid at these amplitudes, we compare our results only with the FD solutions.  We note that nonlinear effects shift the maximum infall velocities toward the density peaks, leading to a slight asymmetry in the velocity profile that is evident in both the GRINN and FD solutions.

\begin{figure}
\centering
\includegraphics[scale=.50]{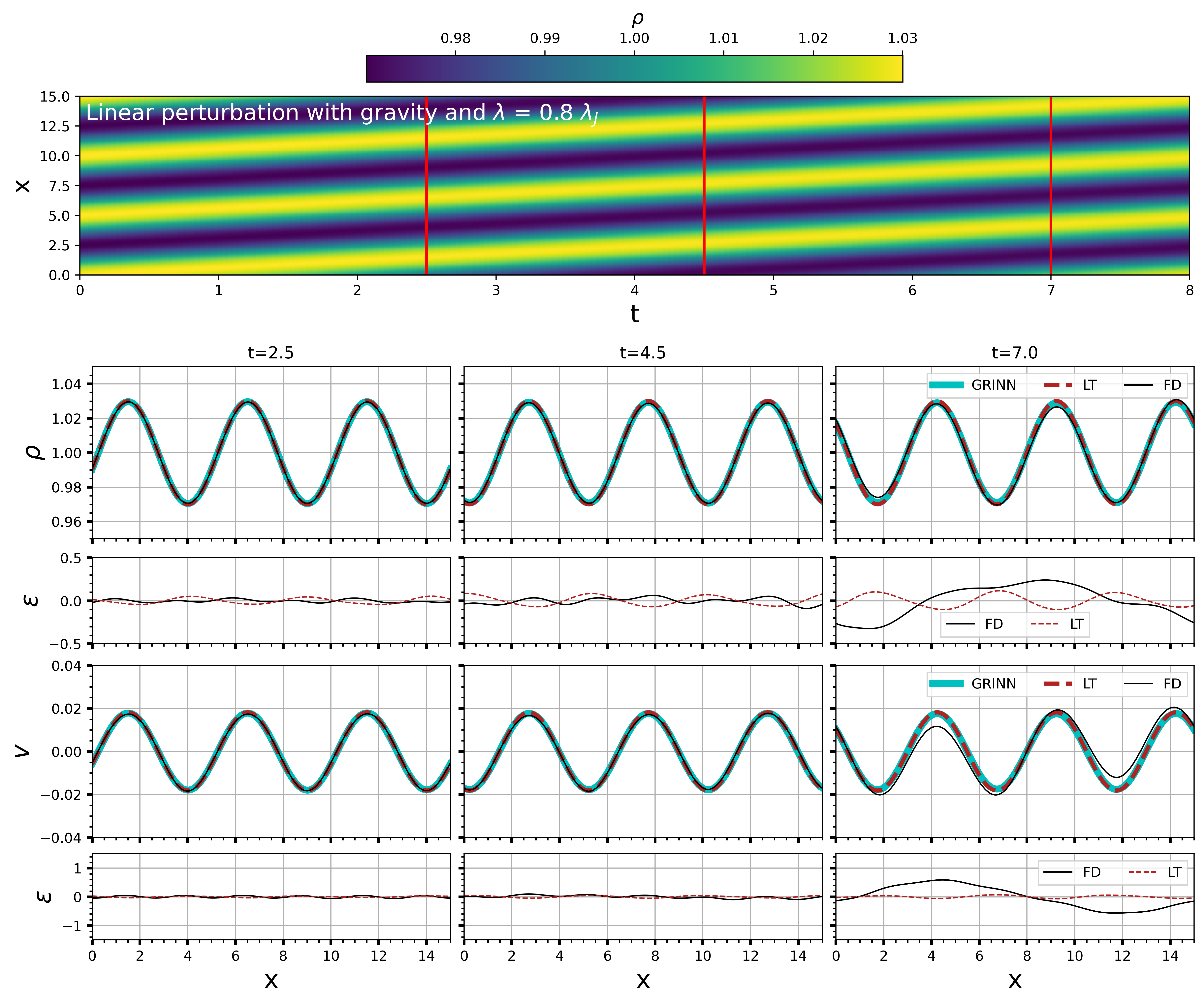}
\caption{Same as Fig.~\ref{fig:jean-linear-gravity} for case~3. The linear perturbation amplitude $\rho_{1a}=0.03$ but wavelength $\lambda = 0.8 \lambda_{J}$, such that the solution consists of sound waves propagating stably but slower than usual due to the extra force of self-gravity.
}\label{fig:NLcorrelation_plot}
\end{figure}

\subsection{Case 3: Propagating linear perturbation with $\lambda<\lambda_{J}$ \label{sec:case3}} 
To explore GRINN's performance on a propagating disturbance, we take as initial condition a linear solution for the density $\rho$ and velocity $\textbf{v}$ using Eq.~(\ref{ic1}) once again.  We let $\rho_{1a} = 0.03$ and $\lambda = 0.8 \lambda_J$, leading to $v_{1a}=0.018$ by Eq.~(\ref{vamp}) in Appendix~\ref{LT}.
For $\lambda<\lambda_{J}$, the linear theory indicates wave propagation with no growth, but at a propagation speed less than the isothermal sound speed owing to self-gravity (Appendix~\ref{LT}).
The third panels of Fig.~\ref{fig:c1_3d}  and Fig.~\ref{fig:c_plot} show the GRINN density solution at $t=2$ for an initial perturbation along the $x$-axis with $k_x=k$ and $k_y=k_z=0$. 
The top panel of Fig.~\ref{fig:NLcorrelation_plot} depicts the time evolution of the gas density till one wave crossing time (i.e., $t=8$). Additionally, we show snapshots of gas density and velocity profile at three epochs in the lower panels for a cross-section taken at $y=z=0.6$.  The FD and LT solutions are overplotted for comparison.  The mismatch $\epsilon$ is less than $0.5\%$ and $1.0\%$ in density and velocity, respectively, at all three times.  The FD solution deviates from the LT and GRINN solutions from about time $t=7$ onward, in the sense that the amplitude decreases slowly due to numerical diffusion.  We adjust the Courant condition to $\nu=0.2$ and increase the number of grid points to $N=8000$ to enhance numerical accuracy. The GRINN solution does not exhibit numerical diffusion and continues to match the linear solution throughout the time period covered.

\section{Discussion} \label{sec:discussion}
Physics-informed neural networks like GRINN open promising new directions for solving 3D astrophysical gas flow problems accurately in a time-efficient way.  PINNs work by approximating functions that are global solutions to the target PDEs, an approach quite different from the local linear or quadratic approximations used in FD methods.  PINN-based models can be built on top of publicly-available neural network modules like TensorFlow\citep {tensorflow2015-whitepaper} and PyTorch \citep{NEURIPS2019_9015}, making them easy to develop and maintain.  Since these public modules are designed to run on GPUs, PINN-based solvers can be implemented on GPU clusters with little effort, easing further enhancements in efficiency.  With the aid of more sophisticated architectures such as XPINNs  \citep{XPINN20} one can run models like GRINN on multiple GPUs, thus making it highly scalable for problems having a large computational domain.


Gravity in many astrophysical systems concentrates mass into localized regions, with corresponding steep density gradients.  For traditional FD solvers, resolving such gradients requires densely-spaced grid points.  The high grid resolution furthermore means small time steps are needed to maintain numerical accuracy and stability. This consequently increases the computational expense.  In contrast, PINNs are mesh-free and capable of resolving wide variations in the computational quantities (e.g., high-density regions), and capture nonlinearities (resulting in  steepening gradients)  with only gradual increases in computational cost.  This gives the PINN-based PDE solvers an edge over traditional methods for modeling self-gravitating systems.

In the following subsections, we discuss some of the pros and cons of PINNs for astrophysical applications, comparing computational costs between the GRINN and FD codes in \S\ref{sec:cost}, examining performance in the nonlinear regime where shocks develop in \S\ref{sec:shocking}, exploring extrapolation of the solution beyond the time domain over which the network is trained in \S\ref{sec:extrapolating}, and discussing some practical issues with the implementation of PINN solvers in \S\ref{sec:limitations}.

\begin{figure}
\centering
\includegraphics[scale=.5]{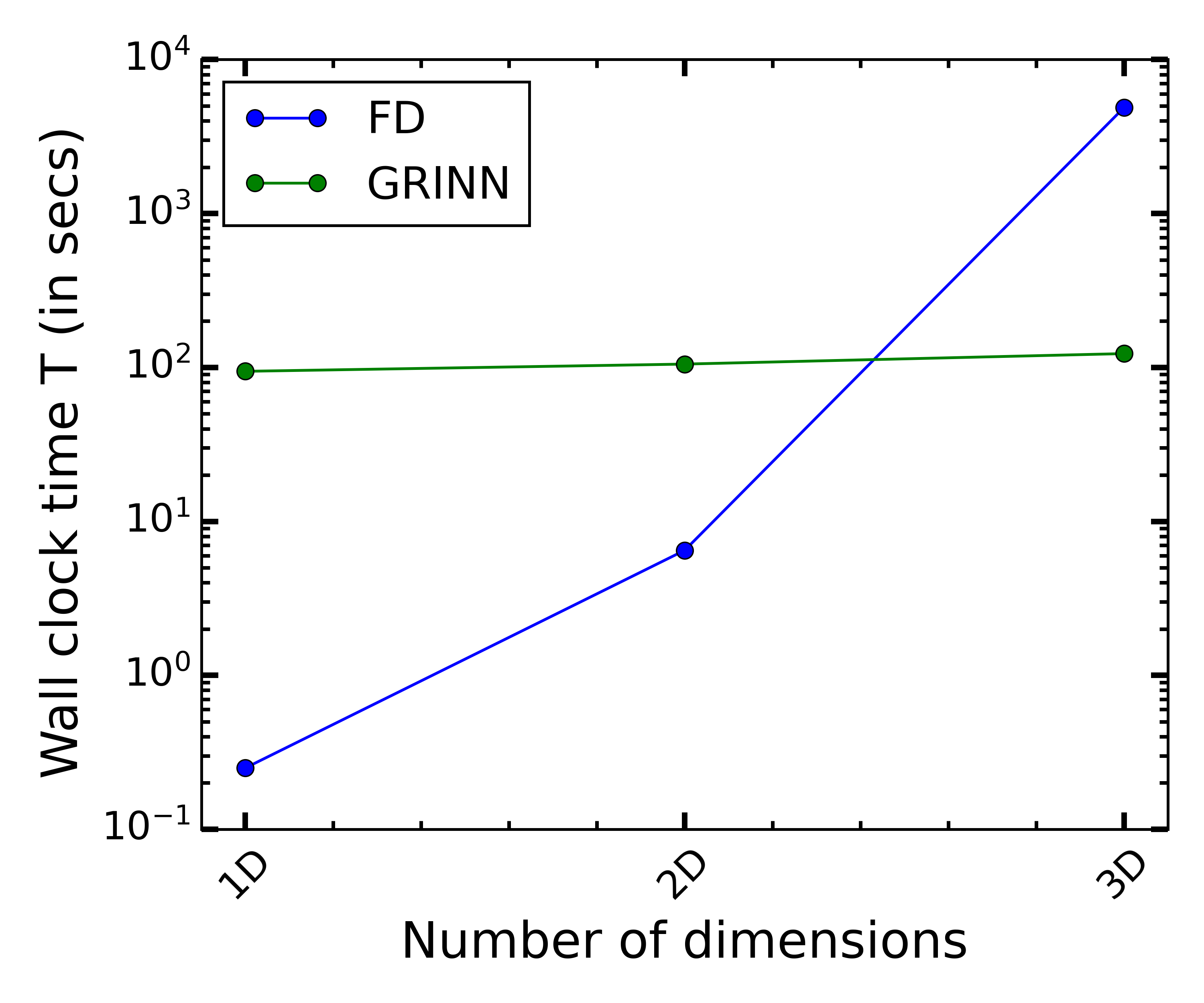}
\includegraphics[scale=.5]{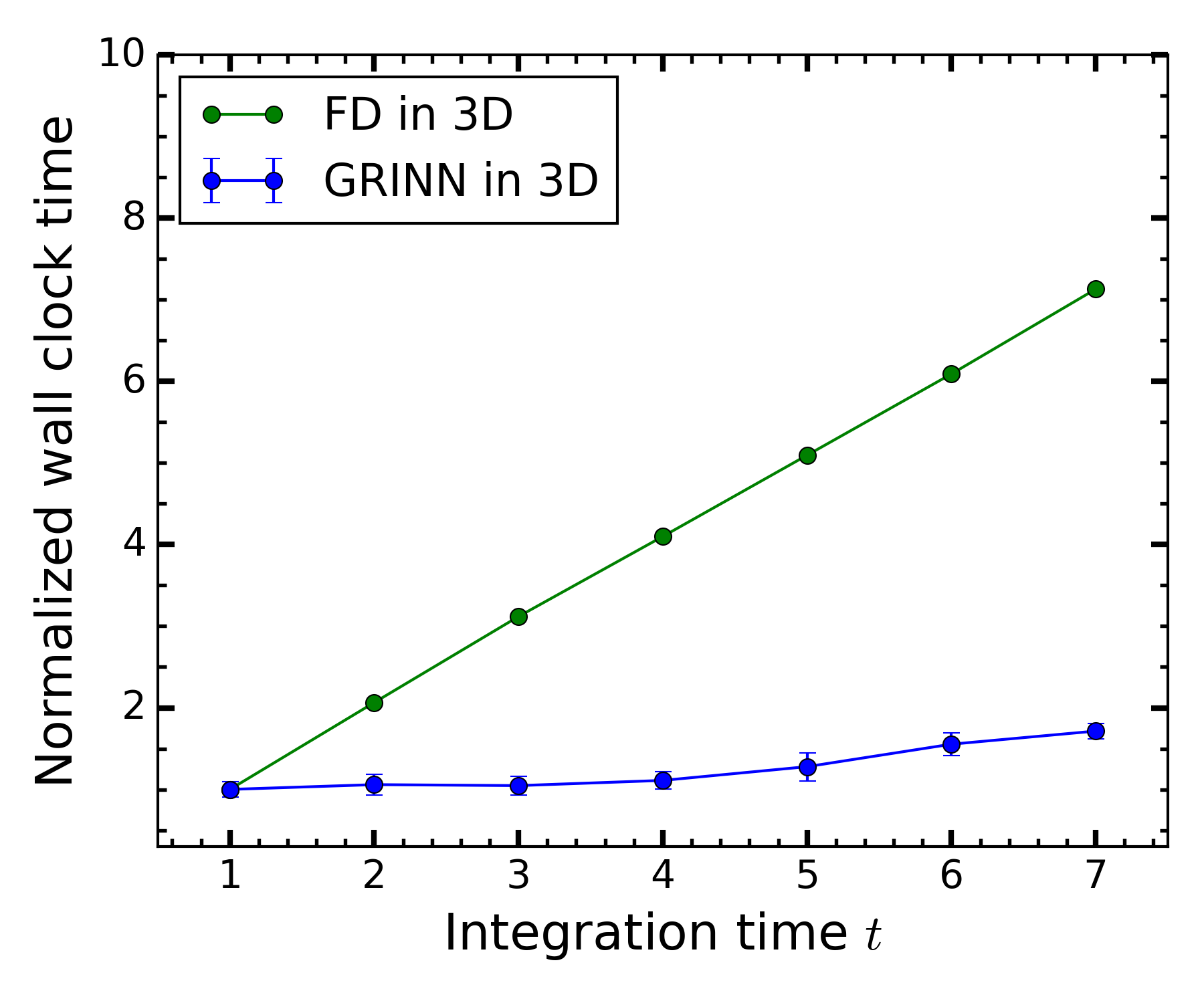}
\caption{Left: Scaling of the run wall clock time with the number of spatial dimensions for the GRINN and FD codes.  Right: Normalized 3D run wall clock times for the two codes vs.\ the problem time interval $t$.  The GRINN model is run twenty times in order to obtain the mean wall clock time and the error bar captures the standard deviation.}\label{fig:scaling}
\end{figure}

\subsection{Computational cost comparison between GRINN and Finite Difference method \label{sec:cost}}
Here we compare the computational cost of the PINN-based hydrodynamic solver against the standard FD code.  We apply both methods to the small and growing perturbations similar to case~1, tracking the  wall clock runtime $T$, for an integration time $t = 7$.  In the FD calculations, we use $N=300$ grid points along each dimension.  We run GRINN on an {\tt\string NVIDIA A100} GPU and the FD calculations on a single core of an {\tt\string INTEL Xeon E7} processor.

Fig. \ref{fig:scaling} shows in the left panel the wall clock time $T$ versus the number $d$ of spatial dimensions over which the hydrodynamic system is solved.  Under the FD scheme, $T$ increases steeply with $d$ due to the geometric rise in the number of grid points $N^d$.  By contrast, GRINN, being mesh-free, has a computational cost almost independent of $d$.  The slight increase in $T$ with $d$ in Fig.~\ref{fig:scaling} is due to the extra operations needed to compute the residual since each added dimension brings an additional equation due to an increase in the components of the velocity (Eq.~\eqref{mom}).  The near-independence of wall clock time on the number of dimensions means that GRINN solves the 3D problem in less than one-tenth the time the FD solver needs.

GRINN also shows better scaling than the FD code with increasing problem integration time $t$.  In the right panel of Fig.~\ref{fig:scaling}, we compare the normalized wall clock times for the same 3D system considered above.  The normalized wall clock time is $T/T_{t=1}$, where $T_{t=1}$ is the computational time for obtaining the solution at $t=1$.  This eliminates the bias induced in the absolute computational time due to external factors such as the choice of hardware and code efficiency. 
For the FD case, the number of time steps $n=t/dt$ (where $dt$ is the discretized time step) increases linearly with the integration time $t$, so the computational time scales as $T\propto t$.  GRINN however being mesh-free solves for the whole space and time domain at once, making the wall clock time almost independent of the integration time. 

\begin{figure}
\centering
\includegraphics[scale=.43]{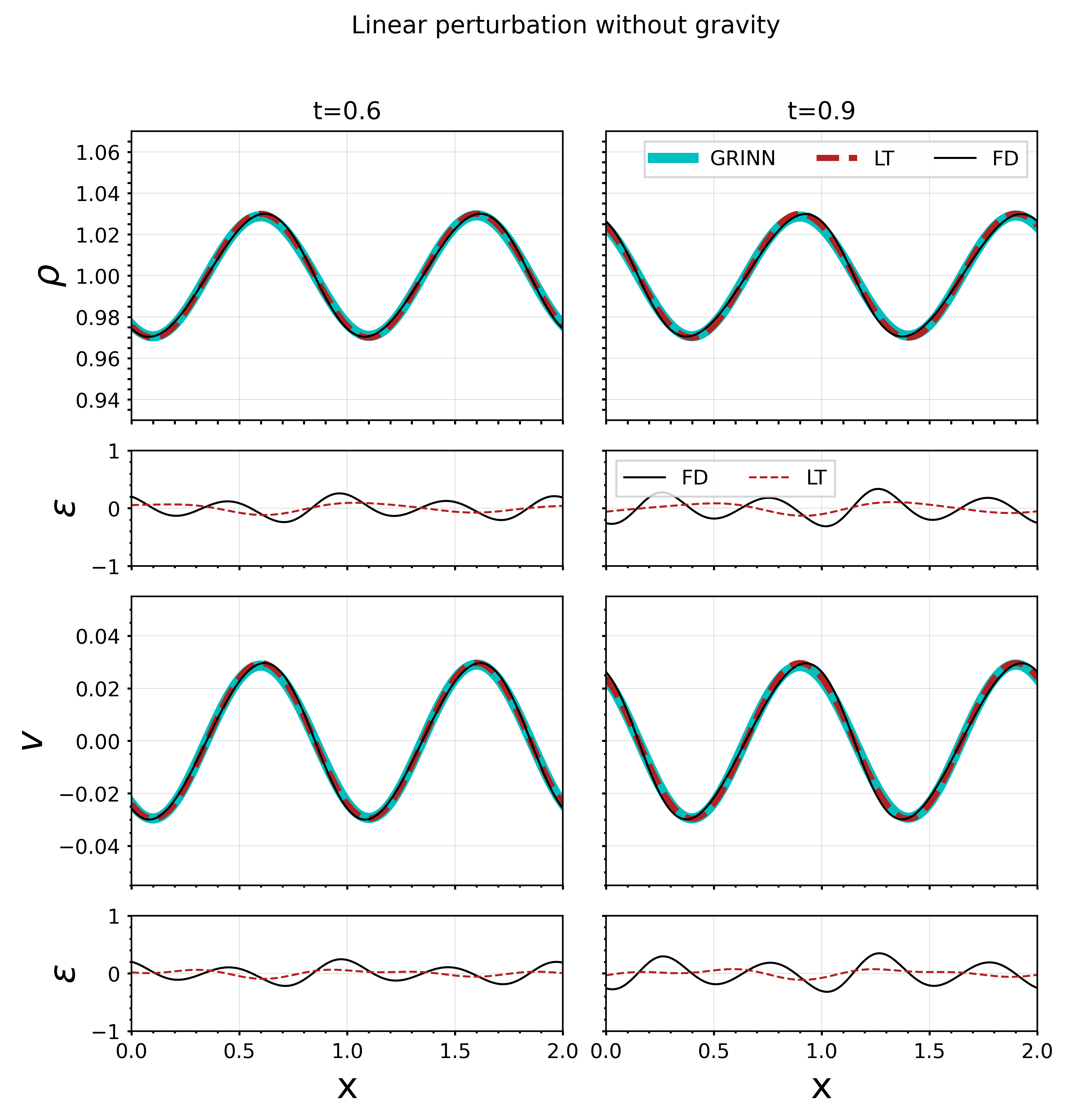}
\includegraphics[scale=.43]{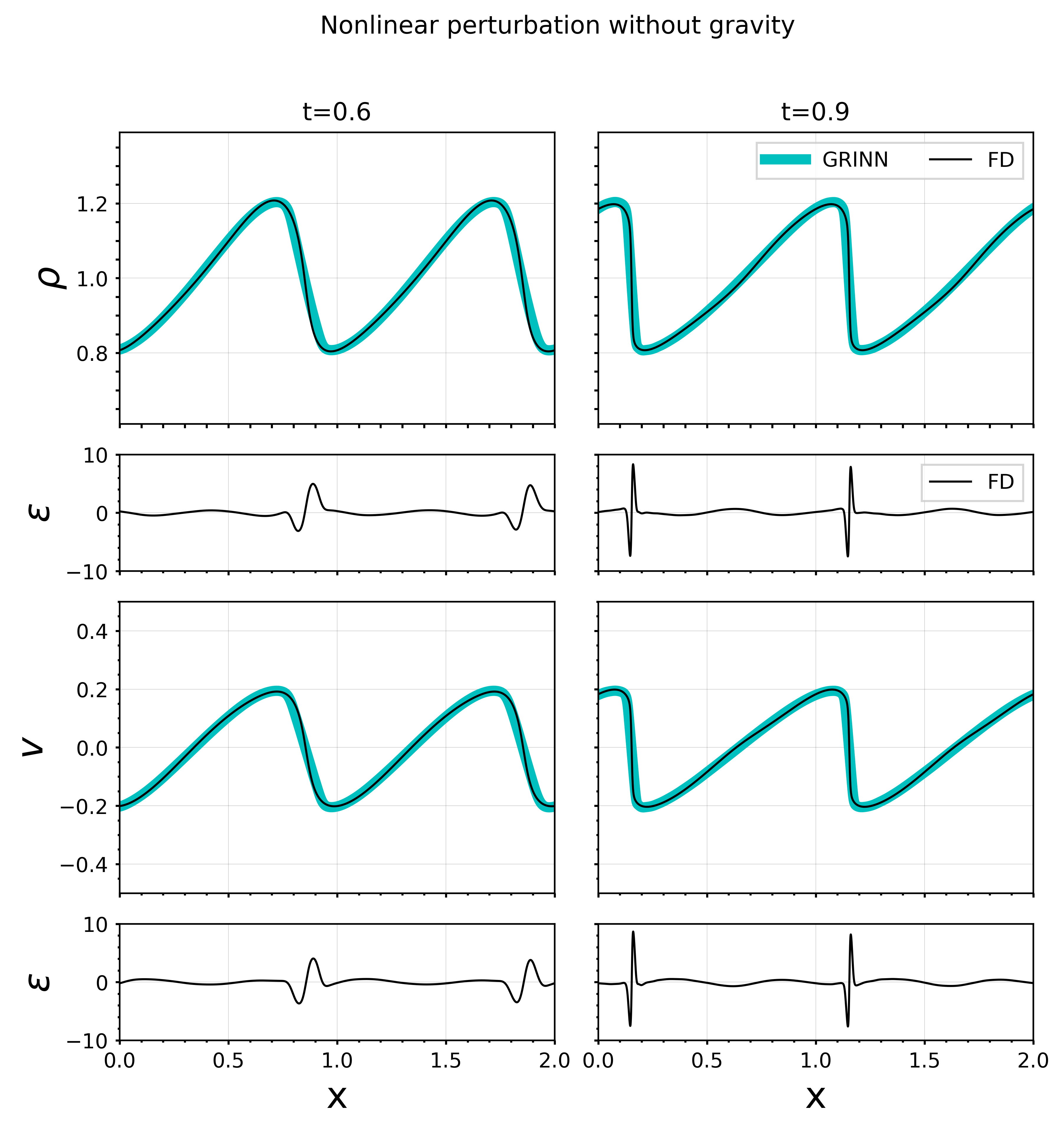}
\caption{Left: GRINN solutions for linear perturbation, $\rho_{1a}=0.03$. The top panel captures the variation in the gas density (at $t=0.6$ and $ t=0.9$) due to the propagation of the sound wave for a non-self gravitating hydrodynamic system. 
The GRINN solution (cyan line) is compared with FD methods (black) and linear theory (LT) (dashed red) results. The last row gives the percentage relative mismatch $\epsilon$ (Eq. \ref{relmisfit}) of the PINNs solution with respect to FD and LT solutions. Right: GRINN solutions for nonlinear perturbation, $\rho_{1a} = 0.2$. The steepening of the wave leads to the development of a shock front. The FD solutions are overplotted for comparison.
}\label{linear_wave}
\end{figure}

\subsection{Sound wave propagation and shock formation in the absence of gravity\label{sec:shocking}}

PINN-based models can solve nonlinear PDEs and capture sharp transitions in the solution such as shocks \cite{coutinho2023physics, MAO2020112789}.
We explore the potential of the GRINN architecture to capture the growth of shocks, considering a nongravitating system. We initialize using the linear solutions of Eqs.~(\ref{rho_init}) and (\ref{v_init}), but without the gravity terms.  We examine both linear and nonlinear initial disturbances, with amplitudes $\rho_{1a} = 0.03$ and  $\rho_{1a} = 0.2$, respectively.  

The two left columns of Fig.~\ref{linear_wave}  show the density and velocity profiles of the propagating sound wave with linear amplitude ($\rho_{1a} = 0.03$) at the two times $t=0.6$ and $0.9$.  The wave propagates at the speed of sound from left to right.
The GRINN solution remains close to LT throughout, with a mismatch $\epsilon<0.3\%$. The mismatches in $\rho$ and $\textbf{v}$ between the GRINN and FD solutions grow with time and are up to $0.5\%$ at $t=0.9$ due to numerical dissipation in the FD scheme.

Next, we initialize the system with a nonlinear perturbation $\rho_{1a} = 0.2$.  
This leads to nonlinear steepening of the wave towards the formation of a shock front.
The two right columns of Fig. \ref{linear_wave} show the GRINN and FD density and velocity solutions along with the relative mismatch.  We lower the mismatch by using a deeper network than for the three test cases in \S\ref{sec:results}, here consisting of 7~hidden layers each with 32~neurons.  Since the linear theory is invalid in this regime, we compare the GRINN solution only with the FD result.  Overall, GRINN captures the nonlinear solution with high precision except at the shock front where the mismatch approaches $\lesssim 10\%$. 
We expect an increasing difference between the solutions particularly because the FD scheme contains numerical dissipation that will smear out the shock front. The GRINN on the other hand is solving equations that contain no innate viscosity. 

As a practical matter, FD codes employ various shock-capturing algorithms \cite{toro2013} and/or artificial viscosity terms to reliably model a shock front. Similarly, for PINN-based models, shock-capturing techniques such as clustered collocation points \cite{MAO2020112789} and adaptive artificial viscosity \cite{coutinho2023physics} have been shown to improve performance on shocks. This will be a fruitful topic to explore in follow-up work on the details of shock modeling, but is not our main topic of interest in this work.
 
\begin{figure}
\centering
\includegraphics[scale=.42]{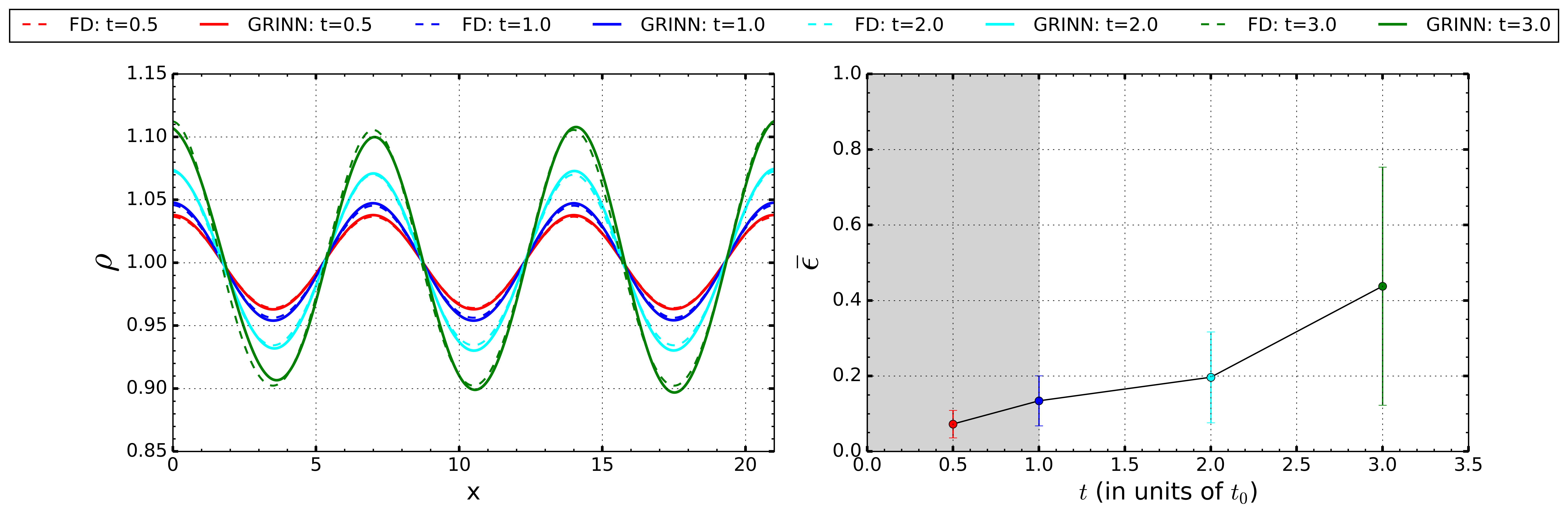}
\includegraphics[scale=.42]{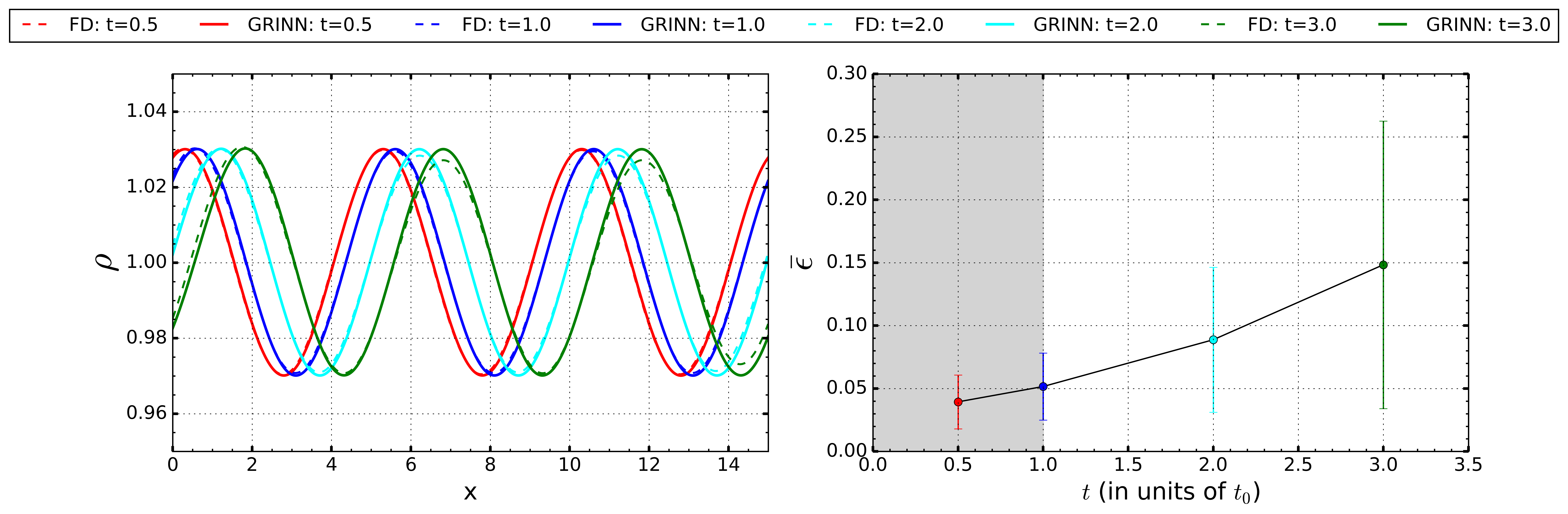}
\caption{Top Left: Extrapolated density solutions (solid colored lines) at four times using GRINN for case~1.  Bottom Left: the same but for case~3.  Dashed colored lines are the corresponding solutions using the FD method. 
Right: Growth of the volume-averaged mismatch $\bar{\epsilon}$ with time in the extrapolated density solutions.  Error bars indicate the domain-wide standard deviation in $\epsilon$. The shaded region marks the time up to which GRINN was trained.}\label{extrapolate}
\end{figure}

\subsection{Extrapolating solutions in time\label{sec:extrapolating}}
In principle, PINNs can extrapolate solutions beyond the time range over which they're trained \cite{kim2021dpm}.  This makes it possible to adapt PINNs for various applications of PDE solutions \cite{zhu2019physics,app11209411,SUN2020112732}).
Extrapolating in time is possible with GRINN as well.  We demonstrate this by obtaining a solution over the interval $t=[0,1]$ for a set of BCs and ICs. Keeping the  parameters $\theta$ of the trained network unchanged, we  apply the network to make predictions up through a later end time $t=3$.  We use this procedure for case~1 and case~3 from \S\ref{sec:results}.  The left top and bottom panels in Fig.~(\ref{extrapolate}) show the density solution versus $x$ at $y=z=0.6$ for case~1 and case~3, respectively.  We compare the solution with the FD result.  The volume-averaged mismatch $\bar{\epsilon}$ remains less than 1\% up to $t=3$.  This shows GRINN can potentially be applied to surrogate modeling \cite{zhu2019physics}, uncertainty quantification, and inverse analysis \cite{2021JCoPh.42509913Y}.

\subsection{Limitations and future scope\label{sec:limitations}}
One of the main limitations of neural network-based architectures is their dependence on the choice of hyperparameters, such as the number of hidden layers, the number of neurons in each layer, and the activation function.  The lack of well-established scaling laws for PINNs makes it difficult to know how to choose these hyperparameters in any given situation.  Most often one resorts to trial-and-error to select the network parameters for specific systems, monitoring the residual and/or the mismatch with known solutions.  Gaining more insight into how each hyperparameter influences the model accuracy and speed will be essential to making PINNs convenient for routine use.

Being a basic deep neural network, GRINN sometimes struggles to converge on an accurate solution as the space and time extent of the domain are increased.  Specifically, GRINN performance suffers when trained on the three test cases out to end times $t>7$.  Adapting advanced techniques like domain decomposition \citep{XPINN20,moseley2021finite} and evolutional deep neural networks \citep{EPINN2021} may help resolve this issue, but this is beyond the scope of our present work.

One can further extend the GRINN framework to build a surrogate model. The idea behind surrogate modeling is to mimic the behavior of a given set of numerical simulations.
Surrogate models can learn the complex relationships between input and output variables for a wide range of parameters describing/defining a physical system.  This is achieved by training  the model with simulations/solutions that span over a finite range of parameters as well as different BCs or ICs. This enables fast and accurate prediction of the system behavior for any of the parameters within the trained domain without explicitly solving the equations or training the model  each time. PINN-based surrogate models can approximate the behavior of complicated physical systems where the underlying dynamics or the physical laws are governed by PDEs.  Leveraging the extrapolation capabilities of GRINN, one can build these surrogate models to emulate the solutions in an effective way \cite{zhu2019physics, 2023CGeot.15905472E}.




\section{Summary and Conclusions}\label{sec:conclusion}
We introduced a physics-informed neural network called GRINN for efficiently solving the coupled set of PDEs describing the evolution of self-gravitating flows in one, two, and three spatial dimensions.  To our knowledge, this is the first demonstration of a PINN for tracing the growth of gravitational instability.  Improved computational speed in modeling such flows could profoundly impact our understanding of star formation, which couples processes operating across a vast range of scales, from the diameter of our Galaxy down to the molecular mean free path that sets the thickness of shocks.  Traditional finite difference and finite volume codes have struggled to span this range, even using adaptive meshes.  PINNs are mesh-free and offer a fundamentally different approach to solving such PDEs.

We investigated three test cases for accuracy and speed relative to a finite difference code that implements the Lax method.  GRINN solved for the evolution of self-gravitating, small-amplitude perturbations about as accurately as the FD code when comparing both against the linear analytic solution.  This was true in the long-wavelength regime, where the perturbations are stationary and unstable with the overdensities collapsing under their own gravity. Further, this holds true in the short-wavelength regime as well, where the perturbations are stable and propagating but travel slower than the regular sound speed owing to their self-gravity.  Finally, long-wavelength unstable perturbations growing into the nonlinear regime and steepening into shocks were evolved similarly in the GRINN and FD codes. 

For the purpose of benchmarking the GRINN code, we ran the test cases in various dimensions.  The wavevector was aligned with one of the grid axes in some calculations and inclined to the axes in others.  GRINN was slower than the FD method in 1D and 2D, owing to the overhead cost of optimizing the network parameters during the training process. This yielded a solution compatible with the PDEs while satisfying the initial and boundary conditions.  
However, the number of operations required for training was almost independent of the number of dimensions involved, in contrast to FD methods where the calculation time increases geometrically with the number of dimensions.  For the 3D calculations, GRINN obtained the final solution more than an order of magnitude faster than the FD code (Fig.~\ref{fig:scaling}).

Our overall result is that the GRINN proved as accurate as the FD method on this set of self-gravitating flow test problems, and significantly faster than the FD method for the subset where the flow was three dimensional. These are largely due to the GRINNs superior scaling as the dimensionality increases and the lack of time step restrictions for stability. We conclude that the GRINN and other physics-informed neural networks hold the potential to substantially increase the scientific community's capability to model the most complex astrophysical flows.

The source code for GRINN is available on the GitHub software repository at \url{https://github.com/sauddy/GRINN}.

\section*{Acknowledgements}
S.A.\ is supported by the NASA Postdoctoral Program (NPP).
S.B.\ is supported by a Discovery Grant from the Natural Sciences and Engineering Research Council of Canada.
This work utilized computing resources provided by the Digital Research Alliance of Canada.
A portion of the research was carried out at the Jet Propulsion Laboratory, California Institute of Technology, under contract 80NM0018D0004 with the National Aeronautics and Space Administration.

\vspace{5mm}






\appendix

\section{Linear Theory}\label{LT}
We linearize the self-gravitating isothermal hydrodynamic Eqs.~(\ref{cont} - \ref{poi}) along lines laid out over a century ago by James Jeans \cite{1902RSPTA.199....1J} over the spatial coordinate $x$, giving the background values subscript $0$ and the small perturbations subscript $1$:
\begin{equation}
    \rho = \rho_0 + \rho_1, v = v_0 + v_1, g = g_0 + g_1,  
\end{equation}
where $v$ is the velocity along the $x$-direction.  We set $v_0 = 0$, such that there is no background velocity. 
We furthermore employ the ``Jeans swindle'' that the background values $g_0$ and $\rho_0$ satisfy Eq.~(\ref{poi}), even though we choose uniform values for $\rho_0$ and $g_0 \,(=0)$.  This inconsistency amounts to assuming that the background state is held in static equilibrium by some unspecified forces.  
Dropping terms above the first order gives  
\begin{eqnarray}
    \frac{\partial \rho_1}{\partial t} & = & - \rho_0 \frac{\partial v_1} {\partial x} \, , \label{linrho}\\
    \rho_0 \frac{\partial v_1}{\partial t}& = &- c_s^{2} \frac{\partial \rho_1}{\partial x}  + \rho_0 g_{1}\, , \\
    \frac{\partial g_1}{\partial x} & = & - 4 \pi G \rho_1\, .
\end{eqnarray}
These three combine into the modified wave equation
\begin{equation}\label{mod_wave}
    \frac{\partial^2 \rho_1}{\partial t ^2} = c_s^{2} \frac{\partial^2 \rho_1}{\partial^2 x} + 4 \pi G \rho_0 \rho_1. 
\end{equation}
A form can also be derived where  $\rho_1$ is eliminated, leaving $v_1$ as the dependent variable.
Since Eq.~(\ref{mod_wave}) and its $v_1$ counterpart have constant coefficients, we can decompose $\rho_1$ and $v_1$ into Fourier components of the form
\begin{equation}
    \rho_1 = \rho_{1a}  e^{i(\omega t - k x)} ,\:  v_1 = v_{1a}  e^{i(\omega t - k x)},
\end{equation}
where $\omega$ and $k$ are the angular frequency and the wavenumber, respectively. We find the dependence of $\omega$ on $k$ by substituting the Fourier components back into Eq.~(\ref{mod_wave}), yielding
\begin{equation}\label{dispersion}
    \omega^2 = c_s^2 k^2 - 4 \pi G \rho_0 \, .
\end{equation}
This dispersion relation indicates instability and exponential growth if $\omega^2 < 0$, or
\begin{eqnarray}
    4\pi G \rho_0 > c_s^2k^2 \implies
    k < k_{J} \equiv \left(\frac{c_s^2}{4 \pi G \rho_0}\right)^{-1/2}.
\end{eqnarray}
Since the perturbation's wavelength $\lambda = 2 \pi / k$, the above implies a special scale, the Jeans length,
\begin{equation}\label{jeans}
     \lambda_{J} \equiv \left(\frac{\pi c_s^2}{G \rho_0}\right)^{1/2},
\end{equation}
such that for $\lambda > \lambda_{J}$ the disturbance grows exponentially, while for $\lambda <\lambda_{J}$ the disturbance is a propagating wave.
Each mode in the physical space is the real part of the complex function, so 
\begin{eqnarray}
    \rho_1 = \Re \left[ \rho_{1a} e^{i(\omega t-kx)}\right] ,\\
    v_1 = \Re \left[ v_{1a} e^{i(\omega t-kx)}\right].
\end{eqnarray}
If $ \rho_{1a}$ is a real number, then 
\begin{eqnarray}\label{rho_init}
      \rho_1 = \Re \left[ \rho_{1a}\left ( \cos (\omega t - kx)+ i \sin (\omega t - kx) \right) \right] 
    = \rho_{1a}\cos(\omega t - kx).
\end{eqnarray}
Similarly for $v_1$, we use Eq. (\ref{linrho}) to get
\begin{eqnarray}\label{v_init}
     &&i \omega \rho_1 - i  k \rho_0 v_1 = 0  \\ \nonumber
     && \implies v_1 = v_{1a} \cos(\omega t - kx) \, ,
\end{eqnarray}
where  $v_{1a} = \frac{\omega}{k}\frac{\rho_{1a}}{\rho_0}$.   
If $\omega$ is real, i.e., $\omega^2 > 0$, we get propagating waves. 
Generally, we can write $\omega = \pm v_p k$, where
\begin{equation}\label{vphase}
v_p = c_s \sqrt{1 - 4 \pi G \rho_0/c_s^2 k^2}
\end{equation}
is the phase speed of the waves. Therefore we can also identify the velocity wave amplitude as
\begin{equation}\label{vamp}
v_{1a} = \pm v_p \frac{\rho_{1a}}{\rho_0} = \pm c_s \frac{\rho_{1a}}{\rho_0} \sqrt{1 - 4 \pi G \rho_0/c_s^2 k^2}\, .
\end{equation}
In the limit of no gravity, $\omega = \pm c_s k$ and  $v_{1a} = \pm c_s \frac{\rho_{1a}}{\rho_0}$ as in ordinary sound waves. 

If $\omega^2 < 0$, then the disturbance is unstable with $\omega = \pm i\alpha$ where
\begin{equation}\label{alpha}
    \alpha = \sqrt{ 4 \pi G \rho_0 - c_s^2 k^2}\,.
\end{equation}
Thus 
\begin{equation}\label{v_pert}
     v_{1a} = \frac{\omega}{k}\frac{\rho_{1a}}{\rho_0} = \frac{\pm i \alpha}{k}\frac{\rho_{1a}}{\rho_0} \,.
\end{equation}
The solution with the minus sign is the exponentially-growing one, having
\begin{equation}\label{vel-per}
v_1 = \Re \left[\frac{- i \alpha}{k}\frac{\rho_{1a}}{\rho_0} e^{i(-i\alpha t - kx)} \right] = - \frac{\alpha}{k}\frac{\rho_{1a}}{\rho_0}e^{\alpha t} \sin(kx)\, .
\end{equation}
Thus the instability's growth time $\tau = \alpha^{-1}$ and in the limit $k\rightarrow 0$ or $\lambda \rightarrow \infty$, $\tau$ approaches the free-fall time $t_0 = 1/\sqrt{4\pi G \rho_0}$.

\section{Finite Difference Method}
For an equation of the form
\begin{equation}
\frac{\partial f}{\partial t} = -\sum_{i=1}^{d}\frac{\partial }{\partial x_{i}}\left(f{\textbf{v}^i}\right) \, ,
\end{equation}
the finite-difference approximation using the Lax method in 3D is
\begin{align}
&\frac{f^{n+1}_{i,j,k}-\left( f^{n}_{i+1,j,k} + f^{n}_{i-1,j,k}+f^{n}_{i,j+1,k} + f^{n}_{i,j-1,k} + +f^{n}_{i,j,k+1} + f^{n}_{i,j,k-1}\right)/ 6}{\Delta t}=
\nonumber\\
&-\frac{f^{n}_{i+1,j,k}v^{n}_{x,i+1,j,k}-f^{n}_{i-1,j,k}v^{n}_{x,i-1,j,k}}{2\Delta x} -\frac{f^{n}_{i,j+1,k}v^{n}_{y,i,j+1,k}-f^{n}_{i,j-1,k}v^{n}_{y,i,j-1,k}}{2\Delta y} 
-\frac{f^{n}_{i,j,k+1}v^{n}_{y,i,j,k+1}-f^{n}_{i,j,k-1}v^{n}_{y,i,j,k-1}}{2\Delta z}\, .
\end{align}
We rewrite the momentum Eq. (\ref{mom}) along the $x$-direction in the flux conservative form as
\begin{eqnarray}
\frac{\partial\left(\rho v_{x}\right)}{\partial t} &=&-\frac{\partial}{\partial x}\left(\rho v_{x} v_{x}\right)-\frac{\partial}{\partial y}(\rho v_{y} v_{x})-\frac{\partial}{\partial z}(\rho v_{z} v_{x}) - c^{2}_{s}\frac{\partial\rho}{\partial x} + \rho\frac{\partial\phi}{\partial x} .
\end{eqnarray}
Similar expressions are also derived along the $y$- and $z$-directions. These three equations along with the continuity equation
\begin{eqnarray}
\frac{\partial \rho}{\partial t} &=& -\frac{\partial}{\partial x}\left(\rho v_{x}\right) - \frac{\partial}{\partial y}\left(\rho v_{y}\right) - \frac{\partial}{\partial z}\left(\rho v_{z}\right)
\end{eqnarray}
are solved using the Lax method. The forward time step is regulated using the Courant condition for stability,
\begin{equation}
\nu \equiv \frac{c\Delta t}{\Delta x} \leq 1\, .
\end{equation}

After each time step, we update the gravitational field $\textbf{g} = - \nabla \phi$ by solving the Poisson equation using a Fast Fourier transform. We solve the Poisson equation on a finite difference grid using second-order differencing. The modes $\phi_{k}$ and $\rho_{k}$ in the wavenumber space are related by
\begin{equation}
\phi_{k} = 2\pi G \rho_{k} 
\left[ \frac{\cos(k_x\Delta x)-1}{\Delta x^2}  + \frac{\cos(k_y\Delta y)-1}{\Delta y^2}  + \frac{\cos(k_z\Delta z)-1}{\Delta z^2} \right]^{-1}\, ,
\end{equation}
where each component $k_{i}= \pm (m\pi)/L$, $m =0,1,2,...,N-1$ and $L$ is the domain length. This relation satisfies the discrete representation of the Poisson equation.

\bibliographystyle{cas-model2-names}

\bibliography{sample631}{}

\begin{thebibliography}{52}
\expandafter\ifx\csname natexlab\endcsname\relax\def\natexlab#1{#1}\fi
\providecommand{\url}[1]{\texttt{#1}}
\providecommand{\href}[2]{#2}
\providecommand{\path}[1]{#1}
\providecommand{\DOIprefix}{doi:}
\providecommand{\ArXivprefix}{arXiv:}
\providecommand{\URLprefix}{URL: }
\providecommand{\Pubmedprefix}{pmid:}
\providecommand{\doi}[1]{\href{http://dx.doi.org/#1}{\path{#1}}}
\providecommand{\Pubmed}[1]{\href{pmid:#1}{\path{#1}}}
\providecommand{\bibinfo}[2]{#2}
\ifx\xfnm\relax \def\xfnm[#1]{\unskip,\space#1}\fi
\bibitem[{Abadi et~al.(2015)Abadi, Agarwal, Barham, Brevdo, Chen, Citro,
  Corrado, Davis, Dean, Devin, Ghemawat, Goodfellow, Harp, Irving, Isard, Jia,
  Jozefowicz, Kaiser, Kudlur, Levenberg, Man\'{e}, Monga, Moore, Murray, Olah,
  Schuster, Shlens, Steiner, Sutskever, Talwar, Tucker, Vanhoucke, Vasudevan,
  Vi\'{e}gas, Vinyals, Warden, Wattenberg, Wicke, Yu and
  Zheng}]{tensorflow2015-whitepaper}
\bibinfo{author}{Abadi, M.}, \bibinfo{author}{Agarwal, A.},
  \bibinfo{author}{Barham, P.}, \bibinfo{author}{Brevdo, E.},
  \bibinfo{author}{Chen, Z.}, \bibinfo{author}{Citro, C.},
  \bibinfo{author}{Corrado, G.S.}, \bibinfo{author}{Davis, A.},
  \bibinfo{author}{Dean, J.}, \bibinfo{author}{Devin, M.},
  \bibinfo{author}{Ghemawat, S.}, \bibinfo{author}{Goodfellow, I.},
  \bibinfo{author}{Harp, A.}, \bibinfo{author}{Irving, G.},
  \bibinfo{author}{Isard, M.}, \bibinfo{author}{Jia, Y.},
  \bibinfo{author}{Jozefowicz, R.}, \bibinfo{author}{Kaiser, L.},
  \bibinfo{author}{Kudlur, M.}, \bibinfo{author}{Levenberg, J.},
  \bibinfo{author}{Man\'{e}, D.}, \bibinfo{author}{Monga, R.},
  \bibinfo{author}{Moore, S.}, \bibinfo{author}{Murray, D.},
  \bibinfo{author}{Olah, C.}, \bibinfo{author}{Schuster, M.},
  \bibinfo{author}{Shlens, J.}, \bibinfo{author}{Steiner, B.},
  \bibinfo{author}{Sutskever, I.}, \bibinfo{author}{Talwar, K.},
  \bibinfo{author}{Tucker, P.}, \bibinfo{author}{Vanhoucke, V.},
  \bibinfo{author}{Vasudevan, V.}, \bibinfo{author}{Vi\'{e}gas, F.},
  \bibinfo{author}{Vinyals, O.}, \bibinfo{author}{Warden, P.},
  \bibinfo{author}{Wattenberg, M.}, \bibinfo{author}{Wicke, M.},
  \bibinfo{author}{Yu, Y.}, \bibinfo{author}{Zheng, X.}, \bibinfo{year}{2015}.
\newblock \bibinfo{title}{{TensorFlow}: Large-scale machine learning on
  heterogeneous systems}.
\newblock \URLprefix \url{https://www.tensorflow.org/}. \bibinfo{note}{software
  available from tensorflow.org}.
\bibitem[{{Auddy} et~al.(2022){Auddy}, {Dey}, {Lin}, {Carrera} and
  {Simon}}]{2022ApJ...936...93A}
\bibinfo{author}{{Auddy}, S.}, \bibinfo{author}{{Dey}, R.},
  \bibinfo{author}{{Lin}, M.K.}, \bibinfo{author}{{Carrera}, D.},
  \bibinfo{author}{{Simon}, J.B.}, \bibinfo{year}{2022}.
\newblock \bibinfo{title}{{Using Bayesian Deep Learning to Infer Planet Mass
  from Gaps in Protoplanetary Disks}}.
\newblock \bibinfo{journal}{ApJ} \bibinfo{volume}{936}, \bibinfo{pages}{93}.
\newblock \DOIprefix\doi{10.3847/1538-4357/ac7a3c},
  \href{http://arxiv.org/abs/2202.11730}{\tt arXiv:2202.11730}.
\bibitem[{{Auddy} et~al.(2021){Auddy}, {Dey}, {Lin} and
  {Hall}}]{2021ApJ...920....3A}
\bibinfo{author}{{Auddy}, S.}, \bibinfo{author}{{Dey}, R.},
  \bibinfo{author}{{Lin}, M.K.}, \bibinfo{author}{{Hall}, C.},
  \bibinfo{year}{2021}.
\newblock \bibinfo{title}{{DPNNet-2.0. I. Finding Hidden Planets from Simulated
  Images of Protoplanetary Disk Gaps}}.
\newblock \bibinfo{journal}{ApJ} \bibinfo{volume}{920}, \bibinfo{pages}{3}.
\newblock \DOIprefix\doi{10.3847/1538-4357/ac1518},
  \href{http://arxiv.org/abs/2107.09086}{\tt arXiv:2107.09086}.
\bibitem[{Baydin et~al.(2018)Baydin, Pearlmutter, Radul and
  Siskind}]{JMLR:v18:17-468}
\bibinfo{author}{Baydin, A.G.}, \bibinfo{author}{Pearlmutter, B.A.},
  \bibinfo{author}{Radul, A.A.}, \bibinfo{author}{Siskind, J.M.},
  \bibinfo{year}{2018}.
\newblock \bibinfo{title}{Automatic differentiation in machine learning: a
  survey}.
\newblock \bibinfo{journal}{Journal of Machine Learning Research}
  \bibinfo{volume}{18}, \bibinfo{pages}{1--43}.
\newblock \URLprefix \url{http://jmlr.org/papers/v18/17-468.html}.
\bibitem[{Berg and Nyström(2018)}]{BERG201828}
\bibinfo{author}{Berg, J.}, \bibinfo{author}{Nyström, K.},
  \bibinfo{year}{2018}.
\newblock \bibinfo{title}{A unified deep artificial neural network approach to
  partial differential equations in complex geometries}.
\newblock \bibinfo{journal}{Neurocomputing} \bibinfo{volume}{317},
  \bibinfo{pages}{28--41}.
\newblock \URLprefix
  \url{https://www.sciencedirect.com/science/article/pii/S092523121830794X},
  \DOIprefix\doi{https://doi.org/10.1016/j.neucom.2018.06.056}.
\bibitem[{Blechschmidt and
  Ernst(2021)}]{https://doi.org/10.1002/gamm.202100006}
\bibinfo{author}{Blechschmidt, J.}, \bibinfo{author}{Ernst, O.G.},
  \bibinfo{year}{2021}.
\newblock \bibinfo{title}{Three ways to solve partial differential equations
  with neural networks — a review}.
\newblock \bibinfo{journal}{GAMM-Mitteilungen} \bibinfo{volume}{44},
  \bibinfo{pages}{e202100006}.
\newblock \URLprefix
  \url{https://onlinelibrary.wiley.com/doi/abs/10.1002/gamm.202100006},
  \DOIprefix\doi{https://doi.org/10.1002/gamm.202100006},
  \href{http://arxiv.org/abs/https://onlinelibrary.wiley.com/doi/pdf/10.1002/gamm.202100006}{\tt
  arXiv:https://onlinelibrary.wiley.com/doi/pdf/10.1002/gamm.202100006}.
\bibitem[{Cai et~al.(2021)Cai, Mao, Wang, Yin and Karniadakis}]{cai2021physics}
\bibinfo{author}{Cai, S.}, \bibinfo{author}{Mao, Z.}, \bibinfo{author}{Wang,
  Z.}, \bibinfo{author}{Yin, M.}, \bibinfo{author}{Karniadakis, G.E.},
  \bibinfo{year}{2021}.
\newblock \bibinfo{title}{Physics-informed neural networks (pinns) for fluid
  mechanics: A review}.
\newblock \bibinfo{journal}{Acta Mechanica Sinica} \bibinfo{volume}{37},
  \bibinfo{pages}{1727--1738}.
\bibitem[{Chen et~al.(2020)Chen, Lu, Karniadakis and
  Dal~Negro}]{chen2020physics}
\bibinfo{author}{Chen, Y.}, \bibinfo{author}{Lu, L.},
  \bibinfo{author}{Karniadakis, G.E.}, \bibinfo{author}{Dal~Negro, L.},
  \bibinfo{year}{2020}.
\newblock \bibinfo{title}{Physics-informed neural networks for inverse problems
  in nano-optics and metamaterials}.
\newblock \bibinfo{journal}{Optics express} \bibinfo{volume}{28},
  \bibinfo{pages}{11618--11633}.
\bibitem[{Courant(1943)}]{bams/1183504922}
\bibinfo{author}{Courant, R.}, \bibinfo{year}{1943}.
\newblock \bibinfo{title}{{Variational methods for the solution of problems of
  equilibrium and vibrations}}.
\newblock \bibinfo{journal}{Bulletin of the American Mathematical Society}
  \bibinfo{volume}{49}, \bibinfo{pages}{1 -- 23}.
\bibitem[{Coutinho et~al.(2023)Coutinho, Dall'Aqua, McClenny, Zhong, Braga-Neto
  and Gildin}]{coutinho2023physics}
\bibinfo{author}{Coutinho, E.J.R.}, \bibinfo{author}{Dall'Aqua, M.},
  \bibinfo{author}{McClenny, L.}, \bibinfo{author}{Zhong, M.},
  \bibinfo{author}{Braga-Neto, U.}, \bibinfo{author}{Gildin, E.},
  \bibinfo{year}{2023}.
\newblock \bibinfo{title}{Physics-informed neural networks with adaptive
  localized artificial viscosity}.
\newblock \bibinfo{journal}{Journal of Computational Physics} ,
  \bibinfo{pages}{112265}.
\bibitem[{Cuoco et~al.(2020)Cuoco, Powell, Cavagli{\`a}, Ackley, Bejger,
  Chatterjee, Coughlin, Coughlin, Easter, Essick et~al.}]{cuoco2020enhancing}
\bibinfo{author}{Cuoco, E.}, \bibinfo{author}{Powell, J.},
  \bibinfo{author}{Cavagli{\`a}, M.}, \bibinfo{author}{Ackley, K.},
  \bibinfo{author}{Bejger, M.}, \bibinfo{author}{Chatterjee, C.},
  \bibinfo{author}{Coughlin, M.}, \bibinfo{author}{Coughlin, S.},
  \bibinfo{author}{Easter, P.}, \bibinfo{author}{Essick, R.}, et~al.,
  \bibinfo{year}{2020}.
\newblock \bibinfo{title}{Enhancing gravitational-wave science with machine
  learning}.
\newblock \bibinfo{journal}{Machine Learning: Science and Technology}
  \bibinfo{volume}{2}, \bibinfo{pages}{011002}.
\bibitem[{Cuomo et~al.(2022)Cuomo, Di~Cola, Giampaolo, Rozza, Raissi and
  Piccialli}]{cuomo2022scientific}
\bibinfo{author}{Cuomo, S.}, \bibinfo{author}{Di~Cola, V.S.},
  \bibinfo{author}{Giampaolo, F.}, \bibinfo{author}{Rozza, G.},
  \bibinfo{author}{Raissi, M.}, \bibinfo{author}{Piccialli, F.},
  \bibinfo{year}{2022}.
\newblock \bibinfo{title}{Scientific machine learning through physics--informed
  neural networks: where we are and what’s next}.
\newblock \bibinfo{journal}{Journal of Scientific Computing}
  \bibinfo{volume}{92}, \bibinfo{pages}{88}.
\bibitem[{D.~Jagtap and Em~Karniadakis(2020)}]{XPINN20}
\bibinfo{author}{D.~Jagtap, A.}, \bibinfo{author}{Em~Karniadakis, G.},
  \bibinfo{year}{2020}.
\newblock \bibinfo{title}{Extended physics-informed neural networks (xpinns): A
  generalized space-time domain decomposition based deep learning framework for
  nonlinear partial differential equations}.
\newblock \bibinfo{journal}{Communications in Computational Physics}
  \bibinfo{volume}{28}, \bibinfo{pages}{2002--2041}.
\newblock \URLprefix
  \url{http://global-sci.org/intro/article_detail/cicp/18403.html},
  \DOIprefix\doi{https://doi.org/10.4208/cicp.OA-2020-0164}.
\bibitem[{Du and Zaki(2021)}]{EPINN2021}
\bibinfo{author}{Du, Y.}, \bibinfo{author}{Zaki, T.A.}, \bibinfo{year}{2021}.
\newblock \bibinfo{title}{Evolutional deep neural network}.
\newblock \bibinfo{journal}{Phys. Rev. E} \bibinfo{volume}{104},
  \bibinfo{pages}{045303}.
\newblock \URLprefix
  \url{https://link.aps.org/doi/10.1103/PhysRevE.104.045303},
  \DOIprefix\doi{10.1103/PhysRevE.104.045303}.
\bibitem[{{Eghbalian} et~al.(2023){Eghbalian}, {Pouragha} and
  {Wan}}]{2023CGeot.15905472E}
\bibinfo{author}{{Eghbalian}, M.}, \bibinfo{author}{{Pouragha}, M.},
  \bibinfo{author}{{Wan}, R.}, \bibinfo{year}{2023}.
\newblock \bibinfo{title}{{A physics-informed deep neural network for surrogate
  modeling in classical elasto-plasticity}}.
\newblock \bibinfo{journal}{Computers and Geotechnics} \bibinfo{volume}{159},
  \bibinfo{pages}{105472}.
\newblock \DOIprefix\doi{10.1016/j.compgeo.2023.105472},
  \href{http://arxiv.org/abs/2204.12088}{\tt arXiv:2204.12088}.
\bibitem[{Eymard et~al.(2000)Eymard, Gallou{\"e}t and
  Herbin}]{Eymard2000FiniteVM}
\bibinfo{author}{Eymard, R.}, \bibinfo{author}{Gallou{\"e}t, T.},
  \bibinfo{author}{Herbin, R.}, \bibinfo{year}{2000}.
\newblock \bibinfo{title}{Finite volume methods}.
\newblock \bibinfo{journal}{Handbook of Numerical Analysis}
  \bibinfo{volume}{7}, \bibinfo{pages}{713--1018}.
\bibitem[{Fujita(2022)}]{9999971}
\bibinfo{author}{Fujita, K.}, \bibinfo{year}{2022}.
\newblock \bibinfo{title}{Physics-informed neural networks with data and
  equation scaling for time domain electromagnetic fields}, in:
  \bibinfo{booktitle}{2022 Asia-Pacific Microwave Conference (APMC)}, pp.
  \bibinfo{pages}{623--625}.
\newblock \DOIprefix\doi{10.23919/APMC55665.2022.9999971}.
\bibitem[{George and Huerta(2018)}]{george2018deep}
\bibinfo{author}{George, D.}, \bibinfo{author}{Huerta, E.A.},
  \bibinfo{year}{2018}.
\newblock \bibinfo{title}{Deep learning for real-time gravitational wave
  detection and parameter estimation: Results with advanced ligo data}.
\newblock \bibinfo{journal}{Physics Letters B} \bibinfo{volume}{778},
  \bibinfo{pages}{64--70}.
\bibitem[{Haghighat et~al.(2021)Haghighat, Raissi, Moure, Gomez and
  Juanes}]{haghighat2021physics}
\bibinfo{author}{Haghighat, E.}, \bibinfo{author}{Raissi, M.},
  \bibinfo{author}{Moure, A.}, \bibinfo{author}{Gomez, H.},
  \bibinfo{author}{Juanes, R.}, \bibinfo{year}{2021}.
\newblock \bibinfo{title}{A physics-informed deep learning framework for
  inversion and surrogate modeling in solid mechanics}.
\newblock \bibinfo{journal}{Computer Methods in Applied Mechanics and
  Engineering} \bibinfo{volume}{379}, \bibinfo{pages}{113741}.
\bibitem[{Hoffer et~al.(2021)Hoffer, Geiger, Ofner and Kern}]{app11209411}
\bibinfo{author}{Hoffer, J.G.}, \bibinfo{author}{Geiger, B.C.},
  \bibinfo{author}{Ofner, P.}, \bibinfo{author}{Kern, R.},
  \bibinfo{year}{2021}.
\newblock \bibinfo{title}{Mesh-free surrogate models for structural mechanic
  fem simulation: A comparative study of approaches}.
\newblock \bibinfo{journal}{Applied Sciences} \bibinfo{volume}{11}.
\newblock \URLprefix \url{https://www.mdpi.com/2076-3417/11/20/9411},
  \DOIprefix\doi{10.3390/app11209411}.
\bibitem[{Hornik et~al.(1989)Hornik, Stinchcombe and
  White}]{Hornik1989MultilayerFN}
\bibinfo{author}{Hornik, K.}, \bibinfo{author}{Stinchcombe, M.B.},
  \bibinfo{author}{White, H.L.}, \bibinfo{year}{1989}.
\newblock \bibinfo{title}{Multilayer feedforward networks are universal
  approximators}.
\newblock \bibinfo{journal}{Neural Networks} \bibinfo{volume}{2},
  \bibinfo{pages}{359--366}.
\bibitem[{Hrennikoff(2021)}]{10.1115/1.4009129}
\bibinfo{author}{Hrennikoff, A.}, \bibinfo{year}{2021}.
\newblock \bibinfo{title}{{Solution of Problems of Elasticity by the Framework
  Method}}.
\newblock \bibinfo{journal}{Journal of Applied Mechanics} \bibinfo{volume}{8},
  \bibinfo{pages}{A169--A175}.
\newblock \URLprefix \url{https://doi.org/10.1115/1.4009129},
  \DOIprefix\doi{10.1115/1.4009129},
  \href{http://arxiv.org/abs/https://asmedigitalcollection.asme.org/appliedmechanics/article-pdf/8/4/A169/6744200/a169\_1.pdf}{\tt
  arXiv:https://asmedigitalcollection.asme.org/appliedmechanics/article-pdf/8/4/A169/6744200/a169\_1.pdf}.
\bibitem[{Iserles(2008)}]{iserles_2008}
\bibinfo{author}{Iserles, A.}, \bibinfo{year}{2008}.
\newblock \bibinfo{title}{A First Course in the Numerical Analysis of
  Differential Equations}.
\newblock Cambridge Texts in Applied Mathematics. \bibinfo{edition}{2} ed.,
  \bibinfo{publisher}{Cambridge University Press}.
\newblock \DOIprefix\doi{10.1017/CBO9780511995569}.
\bibitem[{{Jeans}(1902)}]{1902RSPTA.199....1J}
\bibinfo{author}{{Jeans}, J.H.}, \bibinfo{year}{1902}.
\newblock \bibinfo{title}{{The Stability of a Spherical Nebula}}.
\newblock \bibinfo{journal}{Philosophical Transactions of the Royal Society of
  London Series A} \bibinfo{volume}{199}, \bibinfo{pages}{1--53}.
\newblock \DOIprefix\doi{10.1098/rsta.1902.0012}.
\bibitem[{Kim et~al.(2021)Kim, Lee, Lee, Jhin and Park}]{kim2021dpm}
\bibinfo{author}{Kim, J.}, \bibinfo{author}{Lee, K.}, \bibinfo{author}{Lee,
  D.}, \bibinfo{author}{Jhin, S.Y.}, \bibinfo{author}{Park, N.},
  \bibinfo{year}{2021}.
\newblock \bibinfo{title}{Dpm: a novel training method for physics-informed
  neural networks in extrapolation} \bibinfo{volume}{35},
  \bibinfo{pages}{8146--8154}.
\bibitem[{{Kratter} and {Lodato}(2016)}]{2016ARA&A..54..271K}
\bibinfo{author}{{Kratter}, K.}, \bibinfo{author}{{Lodato}, G.},
  \bibinfo{year}{2016}.
\newblock \bibinfo{title}{{Gravitational Instabilities in Circumstellar
  Disks}}.
\newblock \bibinfo{journal}{araa} \bibinfo{volume}{54},
  \bibinfo{pages}{271--311}.
\newblock \DOIprefix\doi{10.1146/annurev-astro-081915-023307},
  \href{http://arxiv.org/abs/1603.01280}{\tt arXiv:1603.01280}.
\bibitem[{Lax(1954)}]{lax54}
\bibinfo{author}{Lax, P.D.}, \bibinfo{year}{1954}.
\newblock \bibinfo{title}{Weak solutions of nonlinear hyperbolic equations and
  their numerical computation}.
\newblock \bibinfo{journal}{Communications on Pure and Applied Mathematics}
  \bibinfo{volume}{7}, \bibinfo{pages}{159--193}.
\newblock \URLprefix
  \url{https://onlinelibrary.wiley.com/doi/abs/10.1002/cpa.3160070112},
  \DOIprefix\doi{https://doi.org/10.1002/cpa.3160070112},
  \href{http://arxiv.org/abs/https://onlinelibrary.wiley.com/doi/pdf/10.1002/cpa.3160070112}{\tt
  arXiv:https://onlinelibrary.wiley.com/doi/pdf/10.1002/cpa.3160070112}.
\bibitem[{Liu and Nocedal(1989)}]{liu1989limited}
\bibinfo{author}{Liu, D.C.}, \bibinfo{author}{Nocedal, J.},
  \bibinfo{year}{1989}.
\newblock \bibinfo{title}{On the limited memory bfgs method for large scale
  optimization}.
\newblock \bibinfo{journal}{Mathematical programming} \bibinfo{volume}{45},
  \bibinfo{pages}{503--528}.
\bibitem[{Lu et~al.(2021)Lu, Meng, Mao and Karniadakis}]{lu2021deepxde}
\bibinfo{author}{Lu, L.}, \bibinfo{author}{Meng, X.}, \bibinfo{author}{Mao,
  Z.}, \bibinfo{author}{Karniadakis, G.E.}, \bibinfo{year}{2021}.
\newblock \bibinfo{title}{Deepxde: A deep learning library for solving
  differential equations}.
\newblock \bibinfo{journal}{SIAM review} \bibinfo{volume}{63},
  \bibinfo{pages}{208--228}.
\bibitem[{{Machida} and {Basu}(2019)}]{2019ApJ...876..149M}
\bibinfo{author}{{Machida}, M.N.}, \bibinfo{author}{{Basu}, S.},
  \bibinfo{year}{2019}.
\newblock \bibinfo{title}{{The First Two Thousand Years of Star Formation}}.
\newblock \bibinfo{journal}{ApJ} \bibinfo{volume}{876}, \bibinfo{pages}{149}.
\newblock \DOIprefix\doi{10.3847/1538-4357/ab18a7},
  \href{http://arxiv.org/abs/1904.04424}{\tt arXiv:1904.04424}.
\bibitem[{Mao et~al.(2020)Mao, Jagtap and Karniadakis}]{MAO2020112789}
\bibinfo{author}{Mao, Z.}, \bibinfo{author}{Jagtap, A.D.},
  \bibinfo{author}{Karniadakis, G.E.}, \bibinfo{year}{2020}.
\newblock \bibinfo{title}{Physics-informed neural networks for high-speed
  flows}.
\newblock \bibinfo{journal}{Computer Methods in Applied Mechanics and
  Engineering} \bibinfo{volume}{360}, \bibinfo{pages}{112789}.
\newblock \URLprefix
  \url{https://www.sciencedirect.com/science/article/pii/S0045782519306814},
  \DOIprefix\doi{https://doi.org/10.1016/j.cma.2019.112789}.
\bibitem[{{McKee} and {Ostriker}(2007)}]{mckee2007}
\bibinfo{author}{{McKee}, C.F.}, \bibinfo{author}{{Ostriker}, E.C.},
  \bibinfo{year}{2007}.
\newblock \bibinfo{title}{{Theory of Star Formation}}.
\newblock \bibinfo{journal}{araa} \bibinfo{volume}{45},
  \bibinfo{pages}{565--687}.
\newblock \DOIprefix\doi{10.1146/annurev.astro.45.051806.110602},
  \href{http://arxiv.org/abs/0707.3514}{\tt arXiv:0707.3514}.
\bibitem[{{Mestel} and {Spitzer}(1956)}]{mestel1956}
\bibinfo{author}{{Mestel}, L.}, \bibinfo{author}{{Spitzer}, L., J.},
  \bibinfo{year}{1956}.
\newblock \bibinfo{title}{{Star formation in magnetic dust clouds}}.
\newblock \bibinfo{journal}{MNRAS} \bibinfo{volume}{116}, \bibinfo{pages}{503}.
\newblock \DOIprefix\doi{10.1093/mnras/116.5.503}.
\bibitem[{Moseley et~al.(2021)Moseley, Markham and
  Nissen-Meyer}]{moseley2021finite}
\bibinfo{author}{Moseley, B.}, \bibinfo{author}{Markham, A.},
  \bibinfo{author}{Nissen-Meyer, T.}, \bibinfo{year}{2021}.
\newblock \bibinfo{title}{Finite basis physics-informed neural networks
  (fbpinns): a scalable domain decomposition approach for solving differential
  equations}.
\newblock \href{http://arxiv.org/abs/2107.07871}{\tt arXiv:2107.07871}.
\bibitem[{{Mouschovias} and {Ciolek}(1999)}]{mousch1999}
\bibinfo{author}{{Mouschovias}, T.C.}, \bibinfo{author}{{Ciolek}, G.E.},
  \bibinfo{year}{1999}.
\newblock \bibinfo{title}{{Magnetic Fields and Star Formation: A Theory
  Reaching Adulthood}}, in: \bibinfo{editor}{{Lada}, C.J.},
  \bibinfo{editor}{{Kylafis}, N.D.} (Eds.), \bibinfo{booktitle}{The Origin of
  Stars and Planetary Systems}, p. \bibinfo{pages}{305}.
\bibitem[{{Naab} and {Ostriker}(2017)}]{2017ARA&A..55...59N}
\bibinfo{author}{{Naab}, T.}, \bibinfo{author}{{Ostriker}, J.P.},
  \bibinfo{year}{2017}.
\newblock \bibinfo{title}{{Theoretical Challenges in Galaxy Formation}}.
\newblock \bibinfo{journal}{araa} \bibinfo{volume}{55},
  \bibinfo{pages}{59--109}.
\newblock \DOIprefix\doi{10.1146/annurev-astro-081913-040019},
  \href{http://arxiv.org/abs/1612.06891}{\tt arXiv:1612.06891}.
\bibitem[{Noakoasteen et~al.(2020)Noakoasteen, Wang, Peng and
  Christodoulou}]{noakoasteen2020physics}
\bibinfo{author}{Noakoasteen, O.}, \bibinfo{author}{Wang, S.},
  \bibinfo{author}{Peng, Z.}, \bibinfo{author}{Christodoulou, C.},
  \bibinfo{year}{2020}.
\newblock \bibinfo{title}{Physics-informed deep neural networks for transient
  electromagnetic analysis}.
\newblock \bibinfo{journal}{IEEE Open Journal of Antennas and Propagation}
  \bibinfo{volume}{1}, \bibinfo{pages}{404--412}.
\bibitem[{Parascandolo et~al.(2017)Parascandolo, Huttunen and
  Virtanen}]{parascandolo2017taming}
\bibinfo{author}{Parascandolo, G.}, \bibinfo{author}{Huttunen, H.},
  \bibinfo{author}{Virtanen, T.}, \bibinfo{year}{2017}.
\newblock \bibinfo{title}{Taming the waves: sine as activation function in deep
  neural networks}.
\newblock \URLprefix \url{https://openreview.net/forum?id=Sks3zF9eg}.
\bibitem[{Paszke et~al.(2019)Paszke, Gross, Massa, Lerer, Bradbury, Chanan,
  Killeen, Lin, Gimelshein, Antiga, Desmaison, Kopf, Yang, DeVito, Raison,
  Tejani, Chilamkurthy, Steiner, Fang, Bai and Chintala}]{NEURIPS2019_9015}
\bibinfo{author}{Paszke, A.}, \bibinfo{author}{Gross, S.},
  \bibinfo{author}{Massa, F.}, \bibinfo{author}{Lerer, A.},
  \bibinfo{author}{Bradbury, J.}, \bibinfo{author}{Chanan, G.},
  \bibinfo{author}{Killeen, T.}, \bibinfo{author}{Lin, Z.},
  \bibinfo{author}{Gimelshein, N.}, \bibinfo{author}{Antiga, L.},
  \bibinfo{author}{Desmaison, A.}, \bibinfo{author}{Kopf, A.},
  \bibinfo{author}{Yang, E.}, \bibinfo{author}{DeVito, Z.},
  \bibinfo{author}{Raison, M.}, \bibinfo{author}{Tejani, A.},
  \bibinfo{author}{Chilamkurthy, S.}, \bibinfo{author}{Steiner, B.},
  \bibinfo{author}{Fang, L.}, \bibinfo{author}{Bai, J.},
  \bibinfo{author}{Chintala, S.}, \bibinfo{year}{2019}.
\newblock \bibinfo{title}{Pytorch: An imperative style, high-performance deep
  learning library}, in: \bibinfo{booktitle}{Advances in Neural Information
  Processing Systems 32}. \bibinfo{publisher}{Curran Associates, Inc.}, pp.
  \bibinfo{pages}{8024--8035}.
\newblock \URLprefix
  \url{http://papers.neurips.cc/paper/9015-pytorch-an-imperative-style-high-performance-deep-learning-library.pdf}.
\bibitem[{Raissi et~al.(2019)Raissi, Perdikaris and
  Karniadakis}]{RAISSI2019686}
\bibinfo{author}{Raissi, M.}, \bibinfo{author}{Perdikaris, P.},
  \bibinfo{author}{Karniadakis, G.}, \bibinfo{year}{2019}.
\newblock \bibinfo{title}{Physics-informed neural networks: A deep learning
  framework for solving forward and inverse problems involving nonlinear
  partial differential equations}.
\newblock \bibinfo{journal}{Journal of Computational Physics}
  \bibinfo{volume}{378}, \bibinfo{pages}{686--707}.
\newblock \URLprefix
  \url{https://www.sciencedirect.com/science/article/pii/S0021999118307125},
  \DOIprefix\doi{https://doi.org/10.1016/j.jcp.2018.10.045}.
\bibitem[{Sharma et~al.(2023)Sharma, Chung, Akoush and Ihme}]{en16052343}
\bibinfo{author}{Sharma, P.}, \bibinfo{author}{Chung, W.T.},
  \bibinfo{author}{Akoush, B.}, \bibinfo{author}{Ihme, M.},
  \bibinfo{year}{2023}.
\newblock \bibinfo{title}{A review of physics-informed machine learning in
  fluid mechanics}.
\newblock \bibinfo{journal}{Energies} \bibinfo{volume}{16}.
\newblock \URLprefix \url{https://www.mdpi.com/1996-1073/16/5/2343},
  \DOIprefix\doi{10.3390/en16052343}.
\bibitem[{{Shu} et~al.(1987){Shu}, {Adams} and {Lizano}}]{shu1987}
\bibinfo{author}{{Shu}, F.H.}, \bibinfo{author}{{Adams}, F.C.},
  \bibinfo{author}{{Lizano}, S.}, \bibinfo{year}{1987}.
\newblock \bibinfo{title}{{Star formation in molecular clouds: observation and
  theory.}}
\newblock \bibinfo{journal}{araa} \bibinfo{volume}{25},
  \bibinfo{pages}{23--81}.
\newblock \DOIprefix\doi{10.1146/annurev.aa.25.090187.000323}.
\bibitem[{Sitzmann et~al.(2020)Sitzmann, Martel, Bergman, Lindell and
  Wetzstein}]{sitzmann2020implicit}
\bibinfo{author}{Sitzmann, V.}, \bibinfo{author}{Martel, J.},
  \bibinfo{author}{Bergman, A.}, \bibinfo{author}{Lindell, D.},
  \bibinfo{author}{Wetzstein, G.}, \bibinfo{year}{2020}.
\newblock \bibinfo{title}{Implicit neural representations with periodic
  activation functions}.
\newblock \bibinfo{journal}{Advances in Neural Information Processing Systems}
  \bibinfo{volume}{33}, \bibinfo{pages}{7462--7473}.
\bibitem[{{Springel} et~al.(2006){Springel}, {Frenk} and
  {White}}]{springel2006}
\bibinfo{author}{{Springel}, V.}, \bibinfo{author}{{Frenk}, C.S.},
  \bibinfo{author}{{White}, S.D.M.}, \bibinfo{year}{2006}.
\newblock \bibinfo{title}{{The large-scale structure of the Universe}}.
\newblock \bibinfo{journal}{nat} \bibinfo{volume}{440},
  \bibinfo{pages}{1137--1144}.
\newblock \DOIprefix\doi{10.1038/nature04805},
  \href{http://arxiv.org/abs/astro-ph/0604561}{\tt arXiv:astro-ph/0604561}.
\bibitem[{Sun et~al.(2020)Sun, Gao, Pan and Wang}]{SUN2020112732}
\bibinfo{author}{Sun, L.}, \bibinfo{author}{Gao, H.}, \bibinfo{author}{Pan,
  S.}, \bibinfo{author}{Wang, J.X.}, \bibinfo{year}{2020}.
\newblock \bibinfo{title}{Surrogate modeling for fluid flows based on
  physics-constrained deep learning without simulation data}.
\newblock \bibinfo{journal}{Computer Methods in Applied Mechanics and
  Engineering} \bibinfo{volume}{361}, \bibinfo{pages}{112732}.
\newblock \URLprefix
  \url{https://www.sciencedirect.com/science/article/pii/S004578251930622X},
  \DOIprefix\doi{https://doi.org/10.1016/j.cma.2019.112732}.
\bibitem[{Toro(2013)}]{toro2013}
\bibinfo{author}{Toro, E.F.}, \bibinfo{year}{2013}.
\newblock \bibinfo{title}{Riemann solvers and numerical methods for fluid
  dynamics: a practical introduction}.
\newblock \bibinfo{publisher}{Springer Science \& Business Media}.
\bibitem[{{Tsukamoto} et~al.(2022){Tsukamoto}, {Maury}, {Commer{\c{c}}on},
  {Alves}, {Cox}, {Sakai}, {Ray}, {Zhao} and {Machida}}]{tsuka2022}
\bibinfo{author}{{Tsukamoto}, Y.}, \bibinfo{author}{{Maury}, A.},
  \bibinfo{author}{{Commer{\c{c}}on}, B.}, \bibinfo{author}{{Alves}, F.O.},
  \bibinfo{author}{{Cox}, E.G.}, \bibinfo{author}{{Sakai}, N.},
  \bibinfo{author}{{Ray}, T.}, \bibinfo{author}{{Zhao}, B.},
  \bibinfo{author}{{Machida}, M.N.}, \bibinfo{year}{2022}.
\newblock \bibinfo{title}{{The role of magnetic fields in the formation of
  protostars, disks, and outflows}}.
\newblock \bibinfo{journal}{arXiv e-prints} ,
  \bibinfo{pages}{arXiv:2209.13765}\DOIprefix\doi{10.48550/arXiv.2209.13765},
  \href{http://arxiv.org/abs/2209.13765}{\tt arXiv:2209.13765}.
\bibitem[{{Vogelsberger} et~al.(2020){Vogelsberger}, {Marinacci}, {Torrey} and
  {Puchwein}}]{2020NatRP...2...42V}
\bibinfo{author}{{Vogelsberger}, M.}, \bibinfo{author}{{Marinacci}, F.},
  \bibinfo{author}{{Torrey}, P.}, \bibinfo{author}{{Puchwein}, E.},
  \bibinfo{year}{2020}.
\newblock \bibinfo{title}{{Cosmological simulations of galaxy formation}}.
\newblock \bibinfo{journal}{Nature Reviews Physics} \bibinfo{volume}{2},
  \bibinfo{pages}{42--66}.
\newblock \DOIprefix\doi{10.1038/s42254-019-0127-2},
  \href{http://arxiv.org/abs/1909.07976}{\tt arXiv:1909.07976}.
\bibitem[{{Vorobyov} and {Basu}(2006)}]{vor06}
\bibinfo{author}{{Vorobyov}, E.I.}, \bibinfo{author}{{Basu}, S.},
  \bibinfo{year}{2006}.
\newblock \bibinfo{title}{{The Burst Mode of Protostellar Accretion}}.
\newblock \bibinfo{journal}{ApJ} \bibinfo{volume}{650},
  \bibinfo{pages}{956--969}.
\newblock \DOIprefix\doi{10.1086/507320},
  \href{http://arxiv.org/abs/astro-ph/0607118}{\tt arXiv:astro-ph/0607118}.
\bibitem[{{Yang} et~al.(2021){Yang}, {Meng} and
  {Karniadakis}}]{2021JCoPh.42509913Y}
\bibinfo{author}{{Yang}, L.}, \bibinfo{author}{{Meng}, X.},
  \bibinfo{author}{{Karniadakis}, G.E.}, \bibinfo{year}{2021}.
\newblock \bibinfo{title}{{B-PINNs: Bayesian physics-informed neural networks
  for forward and inverse PDE problems with noisy data}}.
\newblock \bibinfo{journal}{Journal of Computational Physics}
  \bibinfo{volume}{425}, \bibinfo{pages}{109913}.
\newblock \DOIprefix\doi{10.1016/j.jcp.2020.109913},
  \href{http://arxiv.org/abs/2003.06097}{\tt arXiv:2003.06097}.
\bibitem[{Zhang et~al.(2022)Zhang, Dao, Karniadakis and
  Suresh}]{zhang2022analyses}
\bibinfo{author}{Zhang, E.}, \bibinfo{author}{Dao, M.},
  \bibinfo{author}{Karniadakis, G.E.}, \bibinfo{author}{Suresh, S.},
  \bibinfo{year}{2022}.
\newblock \bibinfo{title}{Analyses of internal structures and defects in
  materials using physics-informed neural networks}.
\newblock \bibinfo{journal}{Science advances} \bibinfo{volume}{8},
  \bibinfo{pages}{eabk0644}.
\bibitem[{Zhu et~al.(2019)Zhu, Zabaras, Koutsourelakis and
  Perdikaris}]{zhu2019physics}
\bibinfo{author}{Zhu, Y.}, \bibinfo{author}{Zabaras, N.},
  \bibinfo{author}{Koutsourelakis, P.S.}, \bibinfo{author}{Perdikaris, P.},
  \bibinfo{year}{2019}.
\newblock \bibinfo{title}{Physics-constrained deep learning for
  high-dimensional surrogate modeling and uncertainty quantification without
  labeled data}.
\newblock \bibinfo{journal}{Journal of Computational Physics}
  \bibinfo{volume}{394}, \bibinfo{pages}{56--81}.

\end{thebibliography}

\end{document}